\documentclass{article}
\usepackage{amsmath}
\usepackage{amsfonts}
\usepackage{amsthm}
\usepackage{amssymb}
\usepackage{bm}
\usepackage{fontawesome5}
\usepackage{pgffor}
\usepackage{multirow}
\usepackage{verbatim}
\usepackage{enumitem}
\usepackage{microtype}
\usepackage{graphicx}
\usepackage{subfigure}
\usepackage{booktabs}
\usepackage{url,hyperref}
\usepackage{caption}

\usepackage{comment}
\usepackage{pifont}
\usepackage[misc]{ifsym}

\usepackage{mathtools}
\newtheorem{thm}{Theorem}
\newtheorem{lem}[thm]{Lemma}
\newtheorem{prop}[thm]{Proposition}

\newtheorem{defn}[thm]{Definition}

\usepackage{algorithm2e}

\usepackage[accepted]{icml2024}

\newcommand\blfootnote[1]{
  \begingroup
  \renewcommand\thefootnote{}\footnote{#1}
  \addtocounter{footnote}{-1}
  \endgroup
}

\newcommand{\ie}{\emph{i.e.}}
\newcommand{\eg}{\emph{e.g.}}

\icmltitlerunning{Nonparametric Teaching of Implicit Neural Representations}

\begin{document}

\twocolumn[
\icmltitle{Nonparametric Teaching of Implicit Neural Representations}

\icmlsetsymbol{equal}{*}

\begin{icmlauthorlist}
\icmlauthor{Chen Zhang}{hku,equal}
\icmlauthor{Steven Tin Sui Luo}{uot,equal}
\icmlauthor{Jason Chun Lok Li}{hku}
\icmlauthor{Yik-Chung Wu}{hku}
\icmlauthor{Ngai Wong}{hku}
\end{icmlauthorlist}

\icmlaffiliation{hku}{Department of Electrical and Electronic Engineering, The University of Hong Kong, HKSAR, China}
\icmlaffiliation{uot}{Department of Computer Science, The University of Toronto, Ontario, Canada}

\icmlcorrespondingauthor{Ngai Wong}{nwong@eee.hku.hk}

\icmlkeywords{ Nonparametric teaching, Machine teaching, Implicit neural representation, Neural field}

\vskip 0.3in
]

\printAffiliationsAndNotice{\icmlEqualContribution}

\begin{abstract}
We investigate the learning of implicit neural representation (INR) using an overparameterized multilayer perceptron (MLP) via a novel nonparametric teaching perspective. The latter offers an efficient example selection framework for teaching nonparametrically defined (viz. non-closed-form) target functions, such as image functions defined by 2D grids of pixels. To address the costly training of INRs, we propose a paradigm called Implicit Neural Teaching (INT) that treats INR learning as a nonparametric teaching problem, where the given signal being fitted serves as the target function. The teacher then selects signal fragments for iterative training of the MLP to achieve fast convergence. By establishing a connection between MLP evolution through parameter-based gradient descent and that of function evolution through functional gradient descent in nonparametric teaching, we show \emph{for the first time} that teaching an overparameterized MLP is consistent with teaching a nonparametric learner. This new discovery readily permits a convenient drop-in of nonparametric teaching algorithms to broadly enhance INR training efficiency, demonstrating 30\%+ training time savings across various input modalities.
\end{abstract}

\section{Introduction}

\begin{figure}[ht]
	\vskip 0in
	\begin{center}
		\centerline{\includegraphics[width=\columnwidth]{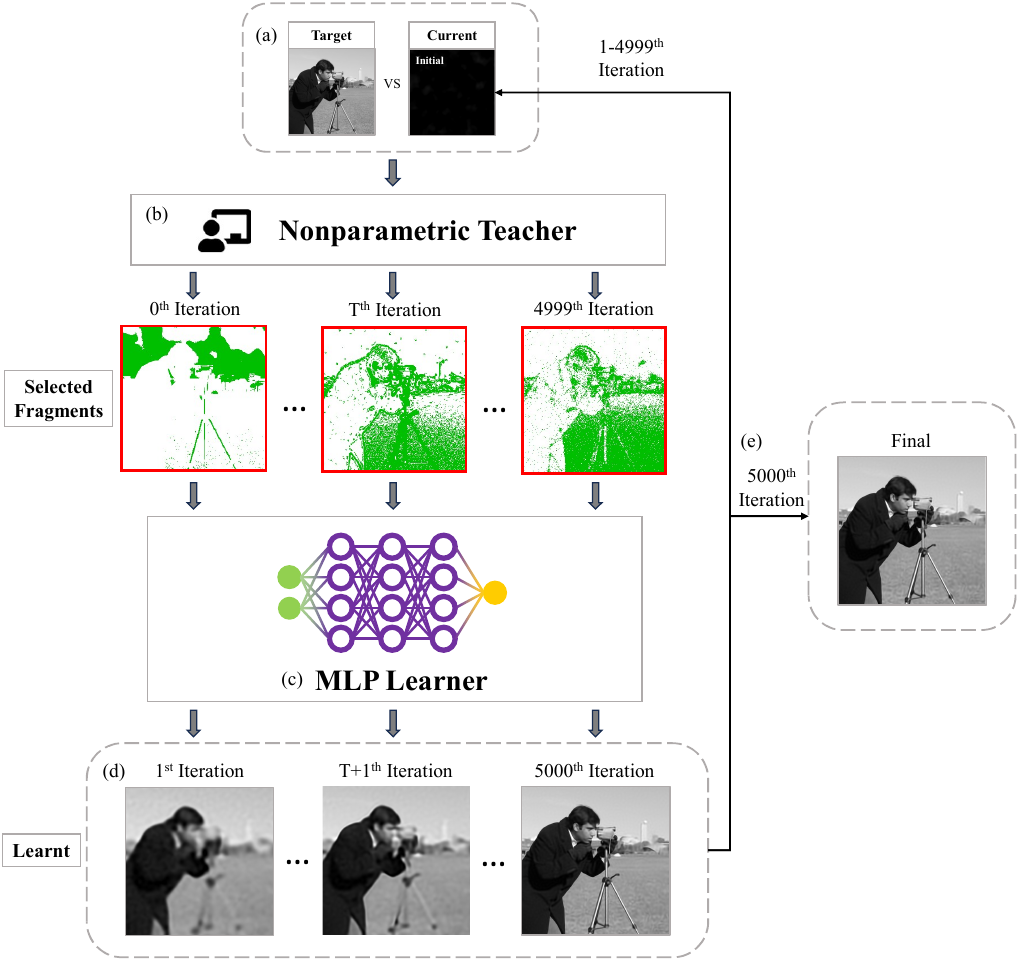}}
		\caption{Fitting a 2D grayscale image signal with Implicit Neural Teaching (INT): By comparing the disparity between the given signal and the current MLP output (a), the nonparametric teacher (b) selectively chooses examples (pixels) of the greatest disparity (red boxes), instead of a raster scan, to feed to the MLP learner (c) who undergoes learning (d) and outputs the final (e).}
		\label{int}
	\end{center}
	\vskip -0.2in
\end{figure}

Implicit neural representation (INR)~\cite{sitzmann2020implicit,tancik2020fourier} focuses on modeling a given signal, which is often discrete, through the use of an overparameterized multilayer perceptron (MLP) such that the signal is accurately fitted by this MLP preserving great details. Such an overparameterized MLP inputs low-dimensional coordinates of the given signal and outputs corresponding values for each input location, \eg, the MLP maps 2D input coordinates to their respective 8-bit levels for a grayscale image. INR has proven to be promising in various domains, including vision data representation~\cite{sitzmann2020implicit,reddy2021multi}, view synthesis~\cite{martin2021nerf,mildenhall2021nerf} and signal compression~\cite{dupont2021coin,pistilli2022signal,strumpler2022implicit,schwarz2023modality}.\blfootnote{Our project page is available at \scriptsize{\url{https://chen2hang.github.io/_publications/nonparametric_teaching_of_implicit_neural_representations/int.html}}.}

Nevertheless, the training of an overparameterized multilayer perceptron (MLP) in INR can be costly, especially when dealing with high-definition signals. For instance, consider the case of a 2D grayscale image with a resolution of $1024\times1024$, which leads to a training set comprising $10^6$ pixels. Additionally, for long videos, the scale of the training set can become prohibitively large. Consequently, it becomes imperative to lower the training cost and enhance the training efficiency of INR.

A recent investigation on nonparametric teaching~\cite{zhang2023nonparametric,zhang2023mint} presents a theoretical framework to facilitate efficient example selection when the target function is nonparametric, \ie, implicitly defined. This inspires a fresh perspective on universally enhancing training efficiency of INR herein. Specifically, machine teaching~\cite{zhu2015machine, liu2017iterative, zhu2018overview} considers the design of a training set (dubbed the teaching set) for the learner, with the goal of enabling speedy convergence towards target functions. Nonparametric teaching~\cite{zhang2023nonparametric,zhang2023mint} relaxes the assumption of target functions being parametric~\cite{liu2017iterative,liu2018towards} to encompass the teaching of nonparametric target functions. In the context of INR, an overparameterized MLP $f$ is akin to a nonparametric function due to its nonlinear activation functions~\cite{leshno1993multilayer} and the inability to be represented solely by its weights $\bm{w}$ as $f(\bm{x})=\langle\bm{w},\bm{x}\rangle$ with input $\bm{x}$~\cite{liu2017iterative,zhang2023nonparametric}, despite appearing to be a parametric function with $\bm{w}$. Unfortunately, the evolution of an MLP is typically achieved by gradient descent on its parameters, whereas nonparametric teaching involves functional gradient descent as the means of function evolution. Bridging this (theoretical + practical) gap is of great value and calls for more examination prior to the application of nonparametric teaching algorithms in the context of INR.

To this end, we recast the evolution achieved through parameter-based gradient descent of an MLP by using dynamic neural tangent kernel (NTK)\footnote{Although NTK for an infinite width MLP remains unchanged during training~\cite{jacot2018neural}, we do not restrict the width of the MLP to be infinite, and instead consider the dynamic NTK.}~\cite{jacot2018neural,lee2019wide,bietti2019inductive,dou2021training}. We express this evolution, from a high-level standpoint of function variation, using functional gradient descent. We show that this dynamic NTK converges to the canonical kernel used in functional gradient descent, indicating that the evolution of the MLP using parameter gradient descent aligns with that using functional gradient descent\footnote{Another example of the alignment is that teaching a parametric function is a special case of nonparametric teaching by using a linear kernel~\cite{zhang2023nonparametric}.}~\cite{geifman2020similarity,chen2020deep}. Therefore, it is natural to cast INR as a nonparametric teaching problem: The given signal to be fitted serves as the target function, and the teacher chooses specific signal fragments prior to providing them to an overparameterized MLP learner, ensuring the MLP fits the signal accurately and efficiently. Consequently, to improve the training efficiency of INR without scenario specification, we propose a novel paradigm called Implicit Neural Teaching (INT), where the teacher leverages the counterpart of the greedy teaching algorithm in nonparametric teaching~\cite{zhang2023nonparametric,zhang2023mint} for INR, namely, selecting examples of the greatest disparity between the given signal and the MLP output~\cite{arbel2019maximum, cormen2022introduction}. Figure~\ref{int} depicts an intuitive illustration of INT. Lastly, we conduct extensive experiments to validate the effectiveness of INT. Our key contributions are:
\begin{itemize}[leftmargin=*,nosep]
	\setlength\itemsep{0.56em}
	\item We propose Implicit Neural Teaching (INT) that novelly interprets implicit neural representation (INR) via the theoretical lens of nonparametric teaching, which in turn enables the utilization of greedy algorithms from the latter to effectively bolster the training efficiency of INRs.
	\item We unveil a strong link between the evolution of a multilayer perceptron (MLP) using gradient descent on its parameters and that of a function using functional gradient descent in nonparametric teaching. This connects nonparametric teaching to MLP training, thus expanding the applicability of nonparametric teaching towards deep learning. We further show that the dynamic NTK, derived from gradient descent on the parameters, converges to the canonical kernel of functional gradient descent.
	\item We showcase the effectiveness of INT through extensive experiments in INR training across multiple modalities. Specifically, INT saves training time  for 1D audio (-31.63\%), 2D images (-38.88\%) and 3D shapes (-35.54\%), while upkeeping its reconstruction quality.
\end{itemize}

\section{Related Works}
\textbf{Implicit neural representation}. There has been a recent surge of interest in implicit neural representation (INR)~\cite{park2019deepsdf,atzmon2020sal,gropp2020implicit,grattarola2022generalised,lindell2022bacon,xie2023diner,li2023regularize,molaei2023implicit,li2023scone, 2024asmr} due to its ability to represent discrete signals continuously. Such representation typically is achieved by training an overparameterized MLP, which offers various practical benefits, including memory efficiency~\cite{sitzmann2020implicit,xie2023diner} and enhanced training efficiency for downstream computer vision tasks~\cite{dupont2022data,chen2023rapid}. There have been various efforts to the accuracy of MLP representation, such as using sinusoidal activation function~\cite{sitzmann2020implicit} and positional encoding with Fourier mapping~\cite{tancik2020fourier}, and to the learning efficiency, such as using a method of dictionary training~\cite{yuce2022structured,wang2022neural} and relying on meta-learning framework~\cite{sitzmann2020metasdf,tancik2021learned, tack2023learning}. Differently, we frame INR from a new perspective as a nonparametric teaching problem~\cite{zhang2023nonparametric,zhang2023mint} and aim to improve the training efficiency by adopting the greedy algorithm from the latter.

\textbf{Nonparametric teaching}. Machine teaching~\cite{zhu2015machine,zhu2018overview} delves into designing a teaching set that leads to a rapid convergence of the learner towards a target model function. It can be seen as an inverse problem of machine learning, in the sense that machine learning aims to learn a function from a given training set while machine teaching aims to construct the set based on a target function. Its applicability has been proven over various domains, such as computer vision~\cite{wang2021gradient,wang2021machine}, robustness~\cite{alfeld2017explicit, ma2019policy, rakhsha2020policy}, and crowd sourcing~\cite{singla2014near, zhou2018unlearn}. Nonparametric teaching~\cite{zhang2023nonparametric,zhang2023mint} improves upon iterative machine teaching~\cite{liu2017iterative,liu2018towards} by extending the parameterized family of target functions to a general nonparametric one. Nevertheless, there are difficulties in directly applying the findings of nonparametric teaching into broadly practical tasks that involves neural networks~\cite{zhang2023nonparametric,zhang2023mint}, which arises due to the gap between nonparametric functions implicitly defined by dense points and overparameterized MLPs. This work bridges this gap using the NTK machinery~\cite{jacot2018neural,lee2019wide,bietti2019inductive,bietti2019kernel,dou2021training}, and shows that teaching an overparameterized MLP is consistent with teaching a nonparametric target function~\cite{gao2019convergence,geifman2020similarity,chen2020deep}. Such insight immediately permits adaptation of tools from the latter to broadly accelerate INR training in the former.

\section{Background}
\textbf{Notation}. To simplify notations, the function being discussed is regarded as scalar-valued without specific emphasis\footnote{In nonparametric teaching, the extension from scalar-valued functions to vector-valued ones, which corresponds to multi-output MLPs, is a well-established generalization in \citealp{zhang2023mint}.}. Let $\mathcal{X}\subseteq\mathbb{R}^n$ denote an $n$ dimensional input (\ie, the coordinate) space and $\mathcal{Y}\subseteq\mathbb{R}$ be an output (\ie, the corresponding value) space. Let a $d$ dimensional column vector with entries $a_i$ indexed by $i\in\mathbb{N}_d$ be $[a_i]_d=(a_1,\cdots,a_d)^T$, where $\mathbb{N}_d\coloneqq\{1,\cdots,d\}$. One may denote it by $\bm{a}$ for simplicity. Likewise, let $\{a_i\}_d$ be a set comprising $d$ elements. Moreover, if the relationship $\{a_i\}_d\subseteq\{a_i\}_n$ is given, then $\{a_i\}_d$ denotes a subset of $\{a_i\}_n$ of size $d$ with the index $i\in\mathbb{N}_n$. By $M_{(i,\cdot)}$ and $M_{(\cdot,j)}$ we refer to the $i$-th row and $j$-th column vector of a matrix $M$, respectively.

Consider $K(\bm{x},\bm{x}'): \mathcal{X}\times\mathcal{X}\mapsto\mathbb{R}$ as a positive definite kernel function. It can be equivalently denoted as $K(\bm{x},\bm{x}')=K_{\bm{x}}(\bm{x}')=K_{\bm{x}'}(\bm{x})$, and $K_{\bm{x}}(\cdot)$ can be shortened as $K_{\bm{x}}$ for brevity. The reproducing kernel Hilbert space (RKHS) $\mathcal{H}$ defined by $K(\bm{x},\bm{x}')$ is the closure of linear span $\{f:f(\cdot)=\sum_{i=1}^{r}a_i K(\bm{x}_i,\cdot),a_i\in\mathbb{R},r\in\mathbb{N},\bm{x}_i\in\mathcal{X}\}$ equipped with inner product $\langle f,g\rangle_\mathcal{H}=\sum_{ij}a_ib_j K(\bm{x}_i,\bm{x}_j)$ when $g=\sum_{j}b_j K_{\bm{x}_j}$~\cite{liu2016stein,arbel2019maximum,shen2020sinkhorn,zhang2023nonparametric}. Given the target signal $f^*:\mathcal{X}\mapsto\mathcal{Y}$, it can uniquely return $y_\dagger$ using the corresponding coordinate $x_\dagger$ as $y_\dagger=f^*(\bm{x}_\dagger)$. By means of the Riesz–Fréchet representation theorem~\cite{lax2002functional, scholkopf2002learning,zhang2023nonparametric}, the evaluation functional is defined as below:
\begin{defn}
	\label{efl}
	For a reproducing kernel Hilbert space $\mathcal{H}$ with the positive definite kernel $K_{\bm{x}}\in\mathcal{H}$ where $\bm{x}\in\mathcal{X}$, the evaluation functional $E_{\bm{x}}(\cdot):\mathcal{H}\mapsto\mathbb{R}$ is defined with the reproducing property as
	\begin{equation}
		E_{\bm{x}}(f)=\langle f, K_{\bm{x}}(\cdot)\rangle_\mathcal{H}=f(\bm{x}),f\in\mathcal{H}.
	\end{equation}
\end{defn}
Furthermore, in the case of a functional $F:\mathcal{H}\mapsto\mathbb{R}$, the Fréchet derivative~\cite{coleman2012calculus, liu2017stein, shen2020sinkhorn,zhang2023nonparametric} of $F$ is presented as follows:
\begin{defn} (Fréchet derivative in RKHS)
	\label{fd}
	The Fréchet derivative of a functional $F:\mathcal{H}\mapsto\mathbb{R}$ at $f\in\mathcal{H}$, which is represented by $\nabla_f F(f)$, is defined implicitly as $F(f+\epsilon g)=F(f)+\langle\nabla_f F(f),\epsilon g\rangle_\mathcal{H}+o(\epsilon)$ for any $g\in\mathcal{H}$ and $\epsilon\in\mathbb{R}$. This derivative is also a function in $\mathcal{H}$.
\end{defn}

\textbf{Nonparametric teaching}. \citealp{zhang2023nonparametric} presents the formulation of nonparametric teaching as a functional minimization over teaching sequence $\mathcal{D}=\{(\bm{x}^1,y^1),\dots(\bm{x}^T,y^T)\}$, with the collection of all possible teaching sequences denoted as $\mathbb{D}$:
\begin{eqnarray}\label{eq1}
	&&\mathcal{D}^*=\underset{\mathcal{D}\in\mathbb{D}}{\arg\min} \mathcal{M}(\hat{f},f^*)+\lambda\cdot \text{len}(\mathcal{D})\nonumber\\
	&&\qquad\text{s.t.}\quad\hat{f}=\mathcal{A}(\mathcal{D}).
\end{eqnarray}
In the above formulation, there are three key elements: $\mathcal{M}$ which measures the disagreement between $\hat{f}$ and $f^*$ (\eg, $L_2$ distance in RKHS $\mathcal{M}(\hat{f^*},f^*)=\|\hat{f^*}-f^*\|_\mathcal{H}$), $\text{len}(\cdot)$ referring to the length of the teaching sequence $\mathcal{D}$ (\ie, the iterative teaching dimension introduced in \citealp{liu2017iterative}) regularized by a constant $\lambda$, and $\mathcal{A}$ which denotes the learning algorithm of learners. Typically, $\mathcal{A}(\mathcal{D})$ employs empirical risk minimization:
\begin{equation}
	\label{la}
	\hat{f}=\underset {f\in\mathcal{H}}{\arg\min}\,\mathbb{E}_{\bm{x}\sim\mathbb{P}(\bm{x})}\left(\mathcal{L}(f(\bm{x}),f^*(\bm{x}))\right)
\end{equation}
with a convex (w.r.t. $f$) loss $\mathcal{L}$, which is optimized by functional gradient descent:
\begin{equation}
	\label{opta}
	f^{t+1}\gets f^t-\eta\mathcal{G}(\mathcal{L},f^*;f^t,\bm{x}^t),
\end{equation}
where $t=0,1,\dots,T$ denotes the time index, $\eta>0$ signifies the learning rate, and $\mathcal{G}$ represents the functional gradient computed at time $t$.

To obtain the functional gradient, which is derived as
\begin{eqnarray}
	\mathcal{G}(\mathcal{L},f^*;f^\dagger,\bm{x})=E_{\bm{x}}\left(\left.\frac{\partial\mathcal{L}(f^*,f)}{\partial f}\right|_{f^\dagger}\right)\cdot K_{\bm{x}},
\end{eqnarray}
\citealp{zhang2023nonparametric,zhang2023mint} introduce the Chain Rule for functional gradients~\cite{gelfand2000calculus} (refer to Lemma~\ref{cr}) and the derivative of the evaluation functional using Fréchet derivative in RKHS~\cite{coleman2012calculus} (cf. Lemma~\ref{ef}).

\begin{lem}(Chain rule for functional gradients)
	\label{cr}
	For differentiable functions $G(F): \mathbb{R}\mapsto\mathbb{R}$ that depends on functionals $F(f):\mathcal{H}\mapsto\mathbb{R}$, the formula 
	\begin{equation}
		\nabla_f G(F(f))=\frac{\partial G(F(f))}{\partial F(f)}\cdot 	\nabla_f F(f)
	\end{equation}
	commonly refers to the chain rule.
\end{lem}
\begin{lem}
	\label{ef}
	The gradient of an evaluation functional $E_{\bm{x}}(f)=f(\bm{x}):\mathcal{H}\mapsto\mathbb{R}$ is $\nabla_f E_{\bm{x}}(f) = K_{\bm{x}}$.
\end{lem}

\section{Implicit Neural Teaching}
We commence by linking the evolution of an MLP that is based on parametric variation with the one that is perceived from a high-level standpoint of function variation. Next, by solving the formulation of MLP evolution as an ordinary differential equation (ODE), we obtain a deeper understanding of this evolution and the underlying cause for its slow convergence. Lastly, we introduce the greedy INT algorithm, which effectively selects examples with steeper gradients at an adaptive batch size and frequency.

\subsection{Evolution of an overparameterized MLP} \label{emlp}
The function represented by an overparameterized MLP $f_\theta\in\mathcal{H}$ with the real-valued parameters $\theta\in\mathbb{R}^m$ (where $m$ denotes the number of parameters in the MLP) is of significant interest~\cite{leshno1993multilayer,gao2019convergence,geifman2020similarity,chen2020deep}. Typically, such an MLP is optimized in terms of a task-specific loss by the method of gradient descent on its parameters~\cite{ruder2016overview}. Given a training set of size $N$ $\{(\bm{x}_i,y_i)|\bm{x}_i\in\mathcal{X},y_i\in\mathcal{Y}\}_N$, the parameter evolves as:
\begin{eqnarray}
	\theta^{t+1}\gets\theta^t-\frac{\eta}{N}\sum_{i=1}^N\nabla_{\theta}\mathcal{L}(f_{\theta^t}(\bm{x}_i),y_i).
\end{eqnarray}
When governed by an extremely small learning rate $\eta$, the update is minute enough over multiple iterations, allowing it to be approximated as a derivative on the time dimension and subsequently transformed into a differential equation:
\begin{eqnarray}
	\frac{\partial\theta^t}{\partial t}=-\frac{\eta}{N}\left[\left.\frac{\partial\mathcal{L}}{\partial f_{\theta}}\right|_{f_{\theta^t},\bm{x}_i}\right]^T_N\cdot \left[\left.\frac{\partial f_{\theta}}{\partial \theta}\right|_{\bm{x}_i,\theta^t}\right]_N.
\end{eqnarray}
Based on Taylor's theorem, it can obtain the evolution of $f_{\theta}$ (a variational representing the variation of $f_{\theta}$ caused by changes in $\theta$) as:
\begin{eqnarray}
	\resizebox{.9\hsize}{!}{$f(\theta^{t+1})-f(\theta^t)=\langle\nabla_{\theta}f(\theta^t),\theta^{t+1}-\theta^t\rangle+o(\theta^{t+1}-\theta^t)$},
\end{eqnarray}
where $f(\theta^\dagger)\coloneqq f_{\theta^\dagger}$. Similar to the transformation of parameter evolution, it can be converted into a differential form in a comparable manner:
\begin{eqnarray} \label{fparat}
	\frac{\partial f_{\theta^t}}{\partial t}= \underbrace{\left\langle\frac{\partial f(\theta^t)}{\partial \theta^t},\frac{\partial \theta^t}{\partial t}\right\rangle}_{(*)} + o\left(\frac{\partial \theta^t}{\partial t}\right).
\end{eqnarray}
It is important to underscore that the nonlinearity of $f(\theta)$ with respect to $\theta$, attributed to the inclusion of nonlinear activation functions, often leads to the remainder $o(\theta^{t+1}-\theta^t)$ not being equal to zero. By substituting the specific parameter evolution into the first-order approximation term $(*)$ of the variational, we obtain
\begin{eqnarray} \label{bfparat}
	\resizebox{.9\hsize}{!}{$\frac{\partial f_{\theta^t}}{\partial t}=-\frac{\eta}{N}\left[\left.\frac{\partial\mathcal{L}}{\partial f_{\theta}}\right|_{f_{\theta^t},\bm{x}_i}\right]^T_N\cdot\left[K_{\theta^t}(\bm{x}_i,\cdot)\right]_N+ o\left(\frac{\partial \theta^t}{\partial t}\right)$},
\end{eqnarray}
where the symmetric and positive definite $K_{\theta^t}(\bm{x}_i,\cdot)=\left\langle\left.\frac{\partial f_{\theta}}{\partial \theta}\right|_{\cdot,\theta^t},\left.\frac{\partial f_{\theta}}{\partial \theta}\right|_{\bm{x}_i,\theta^t} \right\rangle$ (cf. detailed derivation in Appendix~\ref{dpntk}). In a minor distinction, \citealp{jacot2018neural} directly apply the chain rule, paying less heed to the convexity of $\mathcal{L}$ with respect to $\theta$, resulting in the derivation of the first-order approximation as the variational. Meanwhile, $K_{\theta}$ is referred to as the NTK and is demonstrated to remain constant during training by constraining the width of the MLP to be infinite~\cite{jacot2018neural}. In practical terms, it is not necessary for the width of the MLP to be infinitely large, prompting us to explore the dynamic NKT (Appendix~\ref{dpntk} provides an illustration of NTK computation in Figure~\ref{ntk}).

Let the variational be expressed from a high-level standpoint of function variation. Using functional gradient descent,
\begin{eqnarray}
	\frac{\partial f_{\theta^t}}{\partial t}=-\eta\mathcal{G}(\mathcal{L},f^*;f_{\theta^t},\{\bm{x}_i\}_N),
\end{eqnarray}
where the specific functional gradient is 
\begin{eqnarray}
	\resizebox{.9\hsize}{!}{$\mathcal{G}(\mathcal{L},f^*;f_{\theta^t},\{\bm{x}_i\}_N)=\frac{1}{N}\left[\left.\frac{\partial\mathcal{L}}{\partial f_{\theta}}\right|_{f_{\theta^t},\bm{x}_i}\right]^T_N\cdot \left[K({\bm{x}_i},\cdot)\right]_N$}.
\end{eqnarray}
The asymptotic relationship between NTK and the canonical kernel in functional gradient is presented in Theorem~\ref{ntkconverge} below, whose proof is in Appendix~\ref{pntkconverge}.
\begin{thm}\label{ntkconverge}
	For a convex loss $\mathcal{L}$ and a given training set $\{(\bm{x}_i,y_i)|\bm{x}_i\in\mathcal{X},y_i\in\mathcal{Y}\}_N$, the dynamic NTK obtained through gradient descent on the parameters of an overparameterized MLP achieves point-wise convergence to the canonical kernel present in the dual functional gradient with respect to training examples, that is,
	\begin{eqnarray}
		\lim_{t\to\infty}K_{\theta^t}({\bm{x}_i},\cdot)=K({\bm{x}_i},\cdot), \forall i \in\mathbb{N}_N.
	\end{eqnarray}
\end{thm}
It suggests that NTK serves as a dynamic substitute to the canonical kernel used in functional gradient descent, and the evolution of the MLP through parameter gradient descent aligns with that via functional gradient descent~\cite{kuk1995asymptotically,geifman2020similarity,chen2020deep}. This functional insight not only connects the teaching of overparameterized MLPs with that of nonparametric target functions, but also simplifies additional analysis (\eg, a convex functional $\mathcal{L}$ retains the convexity regarding $f_\theta$ in the functional viewpoint, while it is typically nonconvex when considering $\theta$). Through the functional insight and the use of the canonical kernel~\cite{dou2021training} instead of NTK in conjunction with the remainder, it facilitates the derivation of sufficient reduction concerning $\mathcal{L}$ in Proposition~\ref{slr}, with its proof deferred to Appendix~\ref{pslr}.

\begin{prop} (Sufficient Loss Reduction) \label{slr}
	Assuming that the convex loss $\mathcal{L}$ is Lipschitz smooth with a constant $\xi>0$ and the canonical kernel is bounded above by a constant $\zeta>0$, if learning rate $\eta$ satisfies $\eta\leq1/(2\xi\zeta)$, then there exists a sufficient reduction in $\mathcal{L}$ as
	\begin{eqnarray}
		\frac{\partial \mathcal{L}}{\partial t}\leq -\frac{\eta\zeta}{2}\left(\frac{1}{N}\sum_{i=1}^N\left.\frac{\partial\mathcal{L}}{\partial f_{\theta}}\right|_{f_{\theta^t},\bm{x}_i}\right)^2.
	\end{eqnarray}
\end{prop}
It shows that the variation of $\mathcal{L}$ over time is upper bounded by a negative value, which indicates that it decreases by at least the magnitude of this upper bound over time, thereby ensuring convergence.

\subsection{Spectral understanding of the evolution} \label{spu}
\begin{figure}[ht]
	\vskip 0in
	\begin{center}
		\centerline{\includegraphics[width=0.7\columnwidth]{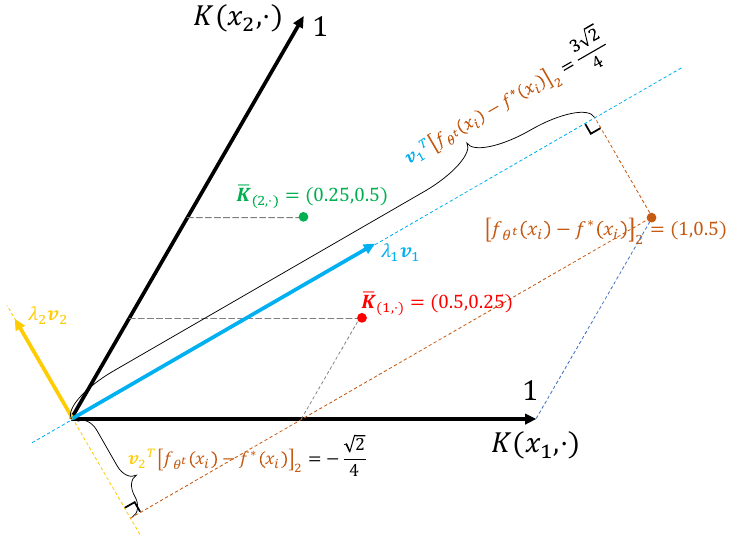}}
		\vskip -0.1in
		\caption{An illustration of the spectral understanding in a 2D function coordinate system (\ie, RKHS) with the $\{K(\bm{x}_i,\cdot)\}_2$ basis. The basis can be non-orthogonal if $K(\bm{x}_i,\bm{x}_j)\neq0$ for $i\neq j$. The coordinate of $f_{\theta^t}-f^*$ represents its projection on each axis, which is given by $\langle\left(f_{\theta^t}-f^*\right),\left[K(\bm{x}_i,\cdot)\right]^T_2\rangle_{\mathcal{H}}=\left[f_{\theta^t}(\bm{x}_i)-f^*(\bm{x}_i)\right]^T_2$, and that of $K(\bm{x}_\dagger,\cdot)$ is $\langle K(\bm{x}_\dagger,\cdot),\left[K(\bm{x}_i,\cdot)\right]^T_2\rangle_{\mathcal{H}}=\left[K(\bm{x}_\dagger,\bm{x}_i)\right]^T_2$, which is stored in the $\dagger$-th row of $\bm{K}$. Assuming $\bar{\bm{K}}=\left[\begin{array}{cc}
				0.5 & 0.25 \\
				0.25 & 0.5 \\
		\end{array}\right]$, the eigenvalues and the respective eigenvectors can be computed as $\lambda_1=0.75,\lambda_2=0.25$ and $\bm{v}_1=(\frac{\sqrt{2}}{2},\frac{\sqrt{2}}{2})^T,\bm{v}_2=(-\frac{\sqrt{2}}{2},\frac{\sqrt{2}}{2})^T$, respectively. Assuming $[f_{\theta^t}(\bm{x}_i)-f^*(\bm{x}_i)]_2$ equals $(1,0.5)$, its first and second principal component projections are $\frac{3\sqrt{2}}{4}$ and $-\frac{\sqrt{2}}{4}$, respectively. Moreover, the discrepancy between $f_{\theta^t}$ and $f^*$ diminishes at a rate of $e^{-\frac{3\eta t}{4}}$ and $e^{-\frac{\eta t}{4}}$ for the first and second principal components, respectively.}
		\label{sde}
	\end{center}
	\vskip -0.3in
\end{figure}

The square loss $\mathcal{L}(f_\theta(\bm{x}),f^*(\bm{x}))=\frac{1}{2}(f_\theta(\bm{x})-f^*(\bm{x}))^2$, commonly used in fitting tasks, is typically used in INR~\cite{sitzmann2020implicit,tancik2020fourier,li2023regularize}. Using this specification for illustration, one obtains the variational of $f_\theta$ from a high-level functional viewpoint:
\begin{eqnarray} \label{fode}
	\frac{\partial f_{\theta^t}}{\partial t}&=&-\eta\mathcal{G}(\mathcal{L},f^*;f_{\theta^t},\{\bm{x}_i\}_N)\nonumber\\
	&=&-\frac{\eta}{N}\left[f_{\theta^t}(\bm{x}_i)-f^*(\bm{x}_i)\right]^T_N\cdot \left[K(\bm{x}_i,\cdot)\right]_N.
\end{eqnarray}
Prior to solving this differential equation, a Lemma of matrix ODE~\cite{godunov1997ordinary,hartman2002ordinary} is in place, with its proof given in Appendix~\ref{pmode}.
\begin{lem} \label{mode}
	Let $\bm{A}$ be an $n\times n$ matrix and $\bm{\alpha}(t)$ be a time-dependent column vector of size $n\times1$. The unique solution of the matrix ODE $\frac{\partial\bm{\alpha}(t)}{\partial t}=\bm{A}\bm{\alpha}(t)$ with initial value $\bm{\alpha}(0)$ is $\bm{\alpha}(t)=e^{\bm{A}t}\bm{\alpha}(0)$, where $e^{\bm{A}t}=\sum_{i=0}^{\infty}\frac{t^i\bm{A}^i}{i!}$.
\end{lem}
Using Lemma~\ref{mode}, Equation~\ref{fode} can be resolved as follows:
\begin{eqnarray}\label{smode}
	\left[f_{\theta^t}(\bm{x}_i)-f^*(\bm{x}_i)\right]_N=e^{-\eta\bar{\bm{K}}t}\cdot\left[f_{\theta^0}(\bm{x}_i)-f^*(\bm{x}_i)\right]_N,
\end{eqnarray}
where $\bar{\bm{K}}=\bm{K}/N$, and $\bm{K}$ is a symmetric and positive definite matrix of size $N\times N$ with entries $K(\bm{x}_i,\bm{x}_j)$ at the $i$-th row and $j$-th column. The comprehensive solution procedure is available in Appendix~\ref{dpode}. Due to the symmetric and positive definite nature of $\bar{\bm{K}}$, it can be orthogonally diagonalized as $\bar{\bm{K}}=\bm{V}\bm{\Lambda} \bm{V}^T$ based on spectral theorem~\cite{hall2013quantum}, where $\bm{V}=[\bm{v}_1,\cdots,\bm{v}_N]$ with column vectors $\bm{v}_i$ representing eigenvectors corresponding to eigenvalue $\lambda_i$, and $\bm{\Lambda}=\text{diag}(\lambda_1,\cdots,\lambda_N)$ is an ordered diagonal matrix ($\lambda_1\geq\cdots\geq\lambda_N$). Hence, we can express $e^{-\eta\bar{\bm{K}}t}$ in a spectral decomposition form as:
\begin{eqnarray}
	e^{-\eta\bar{\bm{K}}t}&=&\bm{I}-\eta t\bm{V}\bm{\Lambda} \bm{V}^T+\frac{1}{2!}\eta^2t^2(\bm{V}\bm{\Lambda} \bm{V}^T)^2+\cdots\nonumber\\
	&=&\bm{V}e^{-\eta\bm{\Lambda} t}\bm{V}^T.
\end{eqnarray}
After rearrangement, Equation~\ref{smode} can be reformulated as:
\begin{eqnarray}\label{pdiff}
	\resizebox{.9\hsize}{!}{$\bm{V}^T\left[f_{\theta^t}(\bm{x}_i)-f^*(\bm{x}_i)\right]_N=\bm{D^t}\bm{V}^T\left[f_{\theta^0}(\bm{x}_i)-f^*(\bm{x}_i)\right]_N$},
\end{eqnarray}
with a diagonal matrix $\bm{D^t}=\text{diag}(e^{-\eta\lambda_1 t},\cdots,e^{-\eta\lambda_N t})$. To be specific, $\left[f_{\theta^0}(\bm{x}_i)-f^*(\bm{x}_i)\right]_N$ refers to the difference vector between $f_{\theta^0}$ and $f^*$ at the initial time, which is evaluated at all training examples, whereas $\left[f_{\theta^t}(\bm{x}_i)-f^*(\bm{x}_i)\right]_N$ denotes the difference vector at time $t$. Additionally, $\bm{V}^T\left[f_{\theta^0}(\bm{x}_i)-f^*(\bm{x}_i)\right]_N$ can be interpreted as the projection of the difference vector onto eigenvectors (\ie, the principal components) at the beginning, while $\bm{V}^T\left[f_{\theta^t}(\bm{x}_i)-f^*(\bm{x}_i)\right]_N$ represents the projection at time $t$. Figure~\ref{sde} provides a lucid illustration in a 2D function coordinate system.

Based on the above, Equation~\ref{pdiff} reveals the connection between the training set and the convergence of $f_{\theta^0}$ towards $f^*$, which indicates that when evaluated on the training set, the discrepancy between $f_{\theta^0}$ and $f^*$ at the $i$-th component exponentially converges to zero at a rate of $e^{-\eta\lambda_i t}$, which is also dependent on the training set~\cite{jacot2018neural}. Meanwhile, this insight uncovers the reason for the sluggish convergence that empirically arises after training for an extended period, wherein small eigenvalues hinder the speed of convergence when continuously training on a static training set. It prompts us to dynamically select examples for fast convergence as described in the next section.

\subsection{INT algorithm}
Intending to make the gradient steeper, the greedy functional teaching algorithm in nonparametric teaching chooses examples by recklessly maximizing the gradient norm: 
\begin{eqnarray}
	{\{\bm{x}_i\}_k}^*=\underset{\{\bm{x}_i\}_k\subseteq\{\bm{x}_i\}_N}{\arg\max}\left\|\mathcal{G}(\mathcal{L},f^*;f_{\theta},\{\bm{x}_i\}_k)\right\|_\mathcal{H},
\end{eqnarray}
where $\mathcal{G}(\mathcal{L},f^*;f_{\theta},\{\bm{x}_i\}_k)=\frac{1}{k}\left[\left.\frac{\partial\mathcal{L}}{\partial f}\right|_{f_{\theta},\bm{x}_i}\right]^T_k\cdot \left[K({\bm{x}_i},\cdot)\right]_k$ and $k\leq N$ denotes the size of selected training set. Drawing from the consistency between an MLP and a nonparametric learner, as explored in Section~\ref{emlp}~\cite{geifman2020similarity,chen2020deep}, we present the INT algorithm that also aims to increase the steepness of gradients. Differently, INT circumvents the potentially cumbersome computation of $\|K(\bm{x}_i,\cdot)\|_\mathcal{H}$ in $\|\mathcal{G}\|_\mathcal{H}$ by utilizing a projection view. To be specific, for $i\in\mathbb{N}_N$, $\frac{\partial\mathcal{L}}{\partial f}|_{f_{\theta},\bm{x}_i}$ can be seen as the component of $\frac{\partial\mathcal{L}}{\partial f}|_{f_\theta}$ projected onto the corresponding element of the basis $\{K(\bm{x}_i,\cdot)\}_N$. Hence, the gradient represents the total sum of the updates, each weighted by $\frac{\partial\mathcal{L}}{\partial f}|_{f_{\theta},\bm{x}_i}$, throughout $\{K(\bm{x}_i,\cdot)\}_k$, which is associated with the selected examples~\cite{wright2015coordinate}. Consequently, steepening the gradient simply requires maximizing the coefficient $\frac{\partial\mathcal{L}}{\partial f}|_{f_{\theta},\bm{x}_i}$, bypassing the need to calculate $\|K(\bm{x}_i,\cdot)\|_\mathcal{H}$. This indicates that selecting examples that enlarge $\left|\frac{\partial\mathcal{L}}{\partial f}|_{f_{\theta},\bm{x}}\right|$ or those which correspond to larger components of $\frac{\partial\mathcal{L}}{\partial f}|_{f_\theta}$ can be sufficient to increase the gradient, which means
\begin{eqnarray}
	{\{\bm{x}_i\}_k}^*=\underset{\{\bm{x}_i\}_k\subseteq\{\bm{x}_i\}_N}{\arg\max}\left\|\left[\left.\frac{\partial\mathcal{L}}{\partial f}\right|_{f_{\theta},\bm{x}_i}\right]_k\right\|_2.
\end{eqnarray}
From a functional perspective, when dealing with a convex loss functional $\mathcal{L}$, the norm of the partial derivative of $\mathcal{L}$ with respect to $f$ at $f_\theta$, denoted as $\|\frac{\partial\mathcal{L}}{\partial f}|_{f_{\theta}}\|_\mathcal{H}$, is positively correlated with$\|f_\theta-f^*\|_\mathcal{H}$; as $f_\theta$ gradually approaches $f^*$, $\|\frac{\partial\mathcal{L}}{\partial f}|_{f_{\theta}}\|_\mathcal{H}$ decrease~\cite{boyd2004convex, coleman2012calculus}. This relationship becomes particularly significant when $\mathcal{L}$ is strongly convex with a larger strong convexity constant~\cite{kakade2008generalization,arjevani2016lower}. Based on these findings, the INT algorithm selects examples by
\begin{eqnarray}\label{intalg}
	{\{\bm{x}_i\}_k}^*=\underset{\{\bm{x}_i\}_k\subseteq\{\bm{x}_i\}_N}{\arg\max}\left\|\left[f_{\theta}(\bm{x}_i)-f^*(\bm{x}_i)\right]_k\right\|_2.
\end{eqnarray}
Pseudo code is in Algorithm~\ref{int_algo}.

\begin{algorithm}[tb]
	\caption{Implicit Neural Teaching}
	\label{int_algo}
	{\bfseries Input:} Target signal $f^*$, initial MLP $f_{\theta^0}$, the size of selected training size $k\leq N$, small constant $\epsilon>0$ and maximal iteration number $T$.
	\BlankLine
	Set $f_{\theta^t}\gets f_{\theta^0}$, $t=0$.
	\BlankLine
	\While{$t\leq T$ {\rm and} $\left\|\left[f_{\theta^t}(\bm{x}_i)-f^*(\bm{x}_i)\right]_N\right\|_2\geq\epsilon$}{
		\BlankLine
		\textbf{The teacher} selects $k$ teaching examples:
            \BlankLine
		\tcc{Examples corresponding to the $k$ largest $|f_{\theta^t}(\bm{x}_i) - f^*(\bm{x}_i)|$.}
  
            ${\{\bm{x}_i\}_k}^*=\underset{\{\bm{x}_i\}_k\subseteq\{\bm{x}_i\}_N}{\arg\max}\left\|\left[f_{\theta^t}(\bm{x}_i)-f^*(\bm{x}_i)\right]_k\right\|_2$.
	
		Provide ${\{\bm{x}_i\}_k}^*$ to the MLP learner.
		\BlankLine
		\textbf{The learner} updates $f_{\theta^t}$ based on received ${\{\bm{x}_i\}_k}^*$:
            \BlankLine
            \tcp{Parameter-based gradient descent.}
		
		$\theta^t\gets \theta^t-\frac{\eta}{k}\sum_{\bm{x}_i\in{\{\bm{x}_i\}_k}^*}\nabla_{\theta}\mathcal{L}(f_{\theta^t}(\bm{x}_i),f^*(\bm{x}_i))$.
  		\BlankLine
		Set $t\gets t+1$.
		}
\end{algorithm}

When considering the square loss commonly employed in INR, the aforementioned correlation can be represented as $\|\frac{\partial\mathcal{L}}{\partial f}|_{f_{\theta}}\|_\mathcal{H}\propto\|f_\theta-f^*\|_\mathcal{H}$.
Besides, it is intriguing that the INT algorithm aligns with the applied variant of the greedy functional teaching algorithm, wherein it is necessary for $\|K(\bm{x}_i,\cdot)\|_\mathcal{H}$ to be uniform or $\|K(\bm{x}_i,\cdot)\|_\mathcal{H}=1$ for all $\bm{x}_i$~\cite{zhang2023nonparametric}. The convergence analysis of the INT algorithm also aligns with that of the greedy functional teaching algorithm obtained in~\citealp{zhang2023nonparametric,zhang2023mint}.

\begin{figure*}[t]
	\subfigbottomskip=-4pt
	\centering
	\includegraphics[width=0.95\linewidth]{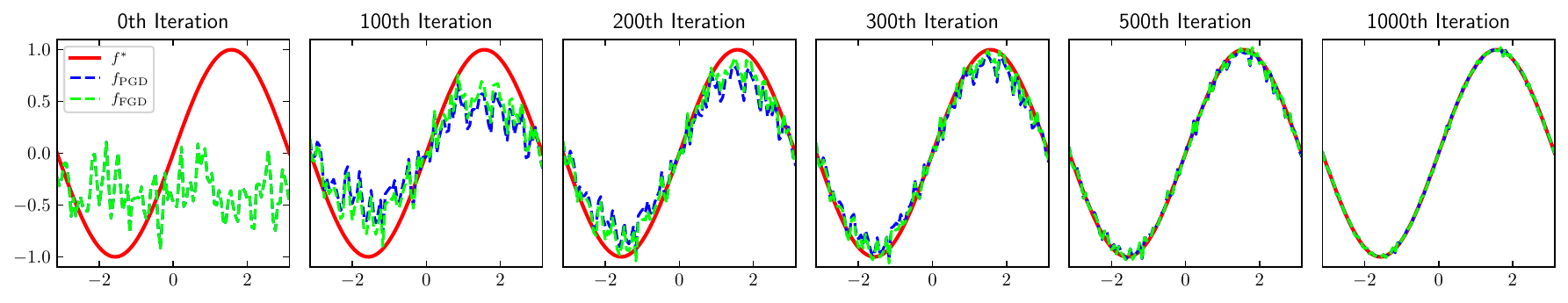}
    \vskip -0.1in
    \caption{Training dynamics of $f$ using PGD and FGD. Apparently, $f_{\text{PGD}}$ closely follows $f_{\text{FGD}}$, empirically showing the evolution consistency between PGD training and FGD training.}
	\label{fig:pgd_fgd}
        \vskip -0.1in
\end{figure*}

With the spectral analysis in Section~\ref{spu}, a deeper understanding of INT follows. First, we define the entire space as the one spanned by the basis corresponding to the whole training set $\{K(\bm{x}_i,\cdot)\}_N$. Similarly, $\{K(\bm{x}_i,\cdot)\}_k\subseteq\{K(\bm{x}_i,\cdot)\}_N$ spans subspaces associated with the selected examples. The eigenvalue of the transformation from the entire space to the subspace of concern (\ie, spanned by $\{K(\bm{x}_i,\cdot)\}_k$ associated with selected examples) is one, while it is zero for the subspace without interest~\cite{watanabe1995discriminative,burgess1996subspace}. The spectral understanding indicates that $f_{\theta^t}$ approaches $f^*$ swiftly at the early stage within the current subspace, owing to the large eigenvalues~\cite{jacot2018neural}. Hence, the INT algorithm can be interpreted as dynamically altering the subspace of interest to fully exploit the period when $f_{\theta^t}$ approaches $f^*$ rapidly. Meanwhile, by selecting examples based on Equation~\ref{intalg}, the subspace of interest is precisely the one where $f_{\theta^t}$ remains significantly distant from $f^*$. In a nutshell, the INT algorithm, by dynamically altering the subspace of interest, not only maximizes the benefits of the fast convergence stage but also updates $f_{\theta^t}$ in the most urgent direction towards $f^*$, thereby saving computational resources compared to training on the entire dataset.

\begin{figure}[t]
    \centering
    \includegraphics[width=1\linewidth]{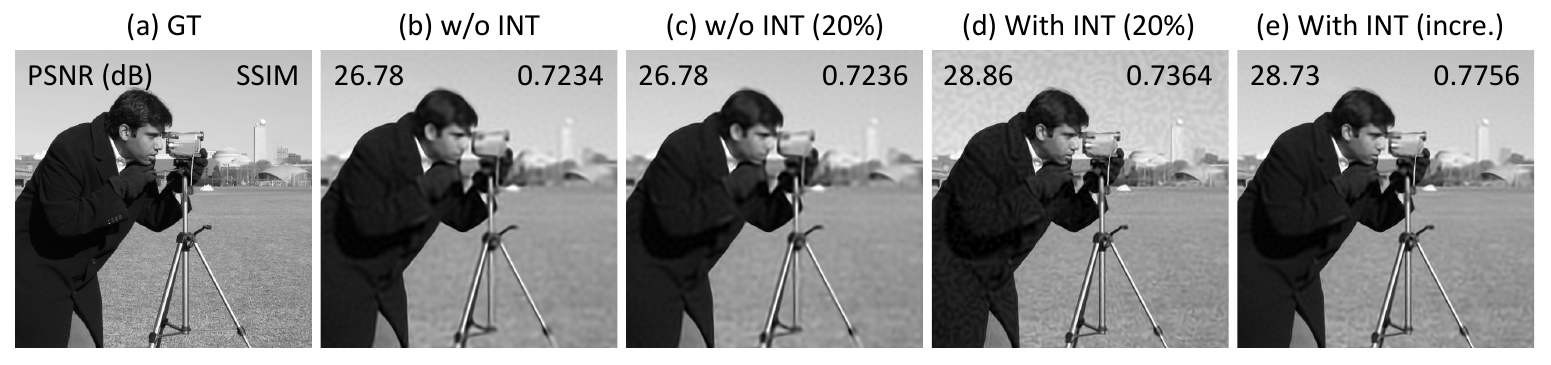}
    \vskip -0.1in
    \caption{Reconstruction quality of SIREN. (b) trains SIREN without (w/o) INT using all pixels. (c) trains it w/o INT using 20\% randomly selected pixels. (d) trains it using INT of 20\% selection rate. (e) trains it using progressive INT (\ie, increasing selection rate progressively from 20\% to 100\%).}
    \label{fig:sgd_pred}
\end{figure}

\section{Experiments and Results}
We begin by using a synthetic signal to empirically show the evolution consistency between parameter-based gradient descent (PGD) and functional gradient descent (FGD). Next, we assess the behavior of INT on a toy image-fitting instance and explore diverse algorithms with different INT frequencies and ratios. Lastly, we validate the INT efficiency in multiple modalities such as audio (-31.63\% training time), images (-38.88\%), and 3D shapes (-35.54\%), while upkeeping its reconstruction quality. Detailed settings are given in Appendices~\ref{ade}.

\paragraph{Synthetic 1D signal.} For an intuitive visualization, we utilize a synthetic 1D signal and present the training dynamics of $f$ obtained through both PGD and FGD. Specifically, the signal (\ie, the target function) is $f^*(x)=\sin(x)$ where $x\in\{x_i\}_{100}$ and is uniformly distributed in the range of $[-\pi, \pi]$. The function corresponding to PGD is obtained by inputting $\{x_i\}_{100}$ into the Fourier Feature network (FFN) trained using PGD, while the function corresponding to FGD is represented by dense points of the nonparametric function updated using FGD. As depicted in Figure~\ref{fig:pgd_fgd}, $f^*$ is well fitted by both PGD and FGD. Moreover, the function obtained through PGD closely mirrors the one obtained through FGD. This observation indicates the consistency in the evolution of the function through both PGD and FGD, suggesting that teaching an overparameterized MLP aligns with teaching a nonparametric target function.

\begin{figure}[t]
    \centering
    \includegraphics[width=\columnwidth]{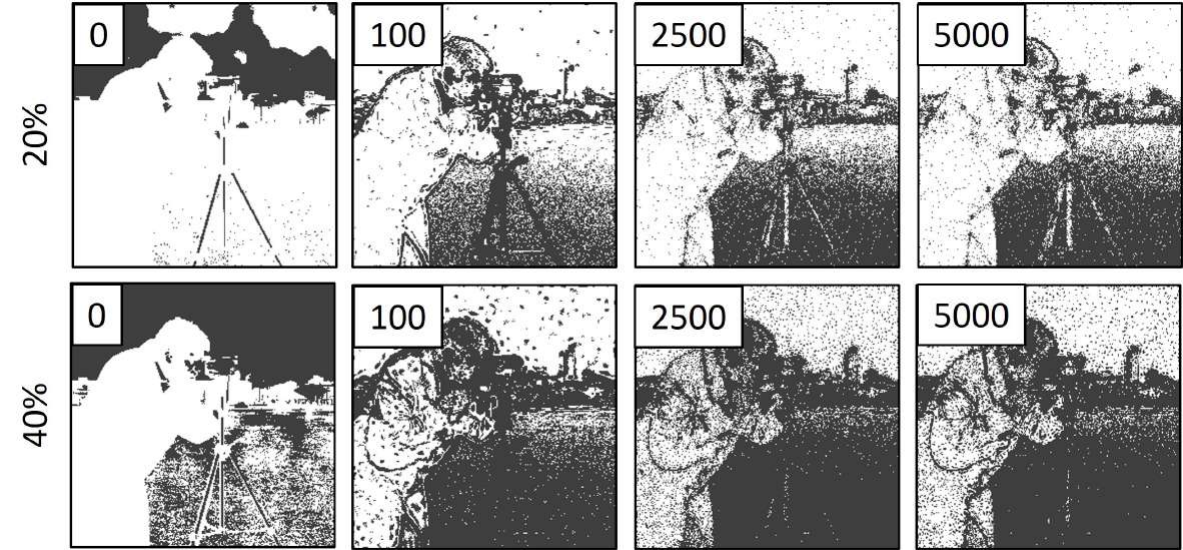}
    \caption{Progression of INT selected pixels (marked as black) at corresponding iterations when training with INT 20\% (top) and 40\% (bottom).}
    \label{fig:mt_dynamics}
\end{figure}

\paragraph{Toy 2D Cameraman fitting.}
In practice, SIREN~\cite{sitzmann2020implicit} is commonly used to encode various modalities of signal such as images. Here, we test the effectiveness of INT in a real-life setting where a SIREN model is used to fit the Cameraman image~\cite{van2014scikit}. We compare the reconstruction quality of SIREN trained with INT of 20\% selection rate (\ie, the size of selected training set is at 20\% of the entire set comprised of all pixels) against that trained without INT, (\ie, using all pixels) and that trained with random sampling at the rate of 20\% at each iteration. INT training results in a higher PSNR and SSIM but exhibits visible artifacts in the background. As shown in Figure~\ref{fig:mt_dynamics} which presents the selected pixels throughout training, we hypothesize that this is due to the over-emphasis of the INT on ``boundary'' pixels where color changes are usually abrupt and hence loss values are larger, leading to an overfitting on the background pixels. On the contrary, using a higher selection rate permits INT to select more examples on the flat surfaces (background), which serves as a regularizer to alleviate the artifacts. Thus, we train an additional SIREN with a progressively increasing INT selection rate from 20\% to nearly 100\%, which achieves superior reconstruction quality without the artifacts.

\begin{figure}[t]
    \centering
    \includegraphics[width=1\linewidth]{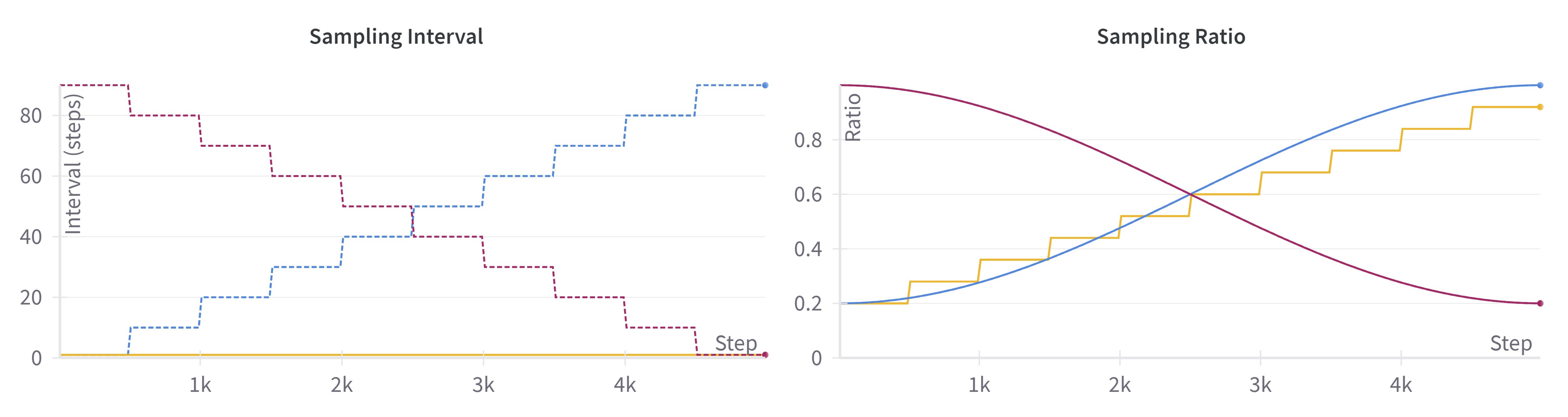}
    \vskip -0.2in
    \caption{Selecting ratio and interval of various INT algorithms. (Left) Red - decremental; Blue - incremental; Yellow - Dense. (Right) Red - R-Cosine; Blue - Cosine; Yellow - Incremental.}
    \label{fig:strategy}
\end{figure}

\begin{table}[t]
    \centering
    \resizebox{\linewidth}{!}{
    \begin{tabular}{cc|cccc}
        \hline
        Ratio & Interval & Time (s) & PSNR (dB)$\uparrow$ & SSIM$\uparrow$ \\
        \hline
        - & - & 345.22 & 35.95$\pm$1.89 & 0.935$\pm$0.03 \\
        Cosine & Dense & 337.00 & 36.39$\pm$2.40 & 0.941$\pm$0.02 \\
        Cosine & Incremental & 227.84 & 36.61$\pm$2.55 & 0.942$\pm$0.02 \\
        R-Cosine & Dense & 346.64 & 35.18$\pm$1.44 & 0.920$\pm$0.02 \\
        R-Cosine & Decremental & 225.30 & 33.56$\pm$2.53 & 0.894$\pm$0.03 \\
        Incremental & Dense & 468.01 & 36.84$\pm$2.70 & \textbf{0.946$\pm$0.02} \\
        \textbf{Incremental} & \textbf{Incremental} & \textbf{211.04} & \textbf{37.04$\pm$2.51} & \textbf{0.946$\pm$0.02} \\
    \bottomrule
    \end{tabular}
    }
    \caption{Performance and training time for different INT strategies on Kodak dataset. The first line (``-'' in both Ratio and Interval) corresponds to training without INT.}
    \label{tab:kodak_strategies}
\end{table}

\paragraph{INT with different frequencies and ratios.}
While using INT can train an INR with fewer examples without sacrificing reconstruction quality, it should be noted that each selecting process requires inferencing all data through the network to rank the difference between the outputs and $f^*$ from higher to lower, which is rather time-consuming. This counters the effect of reducing training time that could originally be brought by the reduction in training examples. Consequently, we follow the observation that increasing the selection ratio leads to increasing overlaps between each consecutive INT selection, and thus devise several INT algorithms that space out the INT frequency (\ie, selecting frequency) and vary the INT ratio (\ie, sizes of selected training sets) dynamically throughout training. Namely, for selecting ratio, we test \textit{constant} ratio, step-wise \textit{increment} of ratio at fixed intervals, and gradually increasing/decreasing the ratio in a \textit{cosine} annealing manner. On the other hand, for sample interval, we test \textit{densely} sampling per iteration, and step-wise \textit{increment}/\textit{decrement} of sampling intervals between 1 and 100 steps.

Figure~\ref{fig:strategy} visualizes the various algorithms we tested against each other. In particular, as presented in Table~\ref{tab:kodak_strategies}, our experiment on a subset of 8 representative images from the Kodak dataset~\cite{kodak} shows that combining an incrementally increasing sampling ratio with an incrementally increasing sampling interval leads to the best performance in terms of both training speed and construction quality. We also want to highlight the severe degradation in reconstruction quality that comes with training an INR via decremental sampling ratio and intervals (comparing rows 4\&5 in Table~\ref{tab:kodak_strategies}). We attribute this to the nature of INRs to progressively learn signals of lower to higher frequencies as shown in~\cite{rahaman2019spectral} while the decremental strategy goes against it. Specifically, at the beginning of training, the MLP may not be able to learn all the information provided by densely sampled examples. But towards the end of training when the MLP is trying to fit the remaining details of the signal, the decremental INT algorithm provides sparser and sparser samples that do not get updated frequently. This serves as a counter-example that explains the effectiveness of utilizing incremental INT for training general INRs, as we shall see in the following section.

\paragraph{INT on multiple real-world modalities.}
To demonstrate the practicality of INT in real-world applications, we conduct experiments on signal fitting tasks across datasets of various modalities, including 1D audio (Librispeech~\cite{panayotov2015librispeech}), 2D images (Kodak~\cite{kodak}), megapixel images (Pluto~\cite{pluto}), and 3D shapes (Stanford 3D Scanning Repository~\cite{stanford3d}). We selected the optimal strategy from Table~\ref{tab:kodak_strategies} (\ie, step-wise increments of both sampling ratio and intervals) as the default INT setting and evaluated it against the baseline without INT. The implementation details of the experiment for each modality can be found in Appendix~\ref{ade}. As shown in Table~\ref{tab:signal_fitting}, it is evident that INT can effectively speed up encoding for all modalities, ranging from 1.41$\times$ to 1.64$\times$, with minimal degradation in performance ($<1$dB PSNR or $<1\%$ IoU). In the case of 2D images, the PSNR with INT even improves from 36.09dB to 36.97dB with near 40\% decrease in training time. We also highlight the results for fitting 3D shapes and megapixel Pluto image (8192$\times$8192), which instead requires mini-batch INT~\cite{zhang2023mint} due to hardware constraints. That is, for each iteration of optimization, we randomly sample a subset of points from the training set and run the INT algorithm to train our model. We make sure that all pixels in the image are sampled for each epoch. This serves as an analogous training procedure to combining stochastic gradient descent with INT and presents the robustness of our INT algorithms in improving training efficiencies.

\begin{table}[t]
\centering
\resizebox{.97\linewidth}{!}{
\begin{tabular}{c|c|cc}
\toprule
    INT                           & Modality & Time (s) & PSNR(dB) / IoU(\%) $\uparrow$ \\
    \midrule
    \multirow{4}{*}{\ding{55}}    & Audio  &  23.05 & 48.38$\pm$3.50 \\
                                  &Image   &  345.22   & 36.09$\pm$2.51 \\
                                  &Megapixel   &  16.78K   & 31.82 \\
                                  &3D Shape&  144.58 & 97.07$\pm$0.84 \\
    \midrule
    \multirow{4}{*}{\checkmark}   &Audio    & 15.76 (-31.63\%) & 48.15$\pm$3.39  \\
                                  &Image   &  211.04 (-38.88\%)  & 36.97$\pm$3.59 \\
                                  &Megapixel   &  11.87K (-29.26\%)   & 33.01 \\
                                  &3D Shape& 93.19 (-35.54\%)& 96.68$\pm$0.83\\
    \bottomrule
    \end{tabular}
    }
\caption{Signal fitting results for different data modalities. The encoding time is measured excluding data I/O latency.}
\label{tab:signal_fitting}
\end{table}

\section{Concluding Remarks and Future Work}
This paper has proposed Implicit Neural Teaching (INT), a novel paradigm that enhances the learning efficiency of implicit neural representation (INR) through nonparametric machine teaching. Using an overparameterized multilayer perceptron (MLP) to fit a given signal, INT reduces the wallclock time for learning INR by over 30\% as demonstrated by extensive experiments. Moreover, INT establishes a theoretically rich connection between the evolution of an MLP using parameter-based gradient descent and that of a function using functional gradient descent in nonparametric teaching. This bridge between nonparametric teaching and MLP training readily expands the applicability of nonparametric teaching in the realm of deep learning. 

Moving forward, it could be more intriguing to explore other practical utilities related to INT towards data efficiency~\cite{henaff2020data, touvron2021training, arandjelovic2021nerf,muller2022instant}. This will involve developing a deeper theoretical understanding of INT, with the neural tangent kernel playing a crucial role. Additionally, exploring more efficient example selection algorithms tailored to specific tasks, such as fine-tuning and prompt training in large language models, holds promise for future advancements.

\section*{Broader Impact} 
Implicit neural representation (INR) has emerged as a promising paradigm in vision data representation, view synthesis and signal compression, domains with significant societal impacts, for its ability of representing discrete signals continuously. This work focuses on enhancing the training efficiency of INR via a novel nonparametric teaching perspective, which can bring positive impacts to INR-related fields and society. 

Meanwhile, this work connects nonparametric teaching to MLP training, which expands the applicability of nonparametric teaching towards deep learning. Thus, it also makes positive contributions to the community of machine teaching. 

Lastly, we are confident that the proposed framework, Implicit Neural Teaching (INT), is highly relevant for enhancing data efficiency and has broader applicability to machine learning tasks, especially in scenarios where the target is known and ``overfitting" is desired, as exhibited in INRs and nonparametric teaching.

\section*{Acknowledgements}
We thank all anonymous reviewers for their constructive feedback to improve our paper. This work was supported by the Theme-based Research Scheme (TRS) project T45-701/22-R, and in part by ACCESS – AI Chip Center for Emerging Smart Systems, sponsored by InnoHK funding, Hong Kong SAR.

\bibliography{main.bib}
\bibliographystyle{icml2024}


\clearpage
\newpage

\appendix
\onecolumn

\begin{appendix}
	
	\thispagestyle{plain}
	\begin{center}
		{\Large \bf Appendix}
	\end{center}
	
\end{appendix}

\section{Additional Discussions}

\begin{figure}[t]
	\vskip 0.2in
	\begin{center}
		\centerline{\includegraphics[width=0.9\columnwidth]{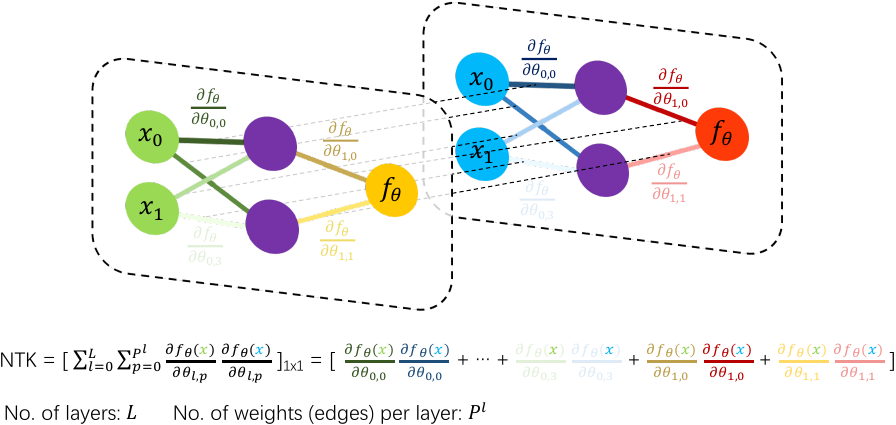}}
		\caption{Graphical illustration of NTK computation.}
		\label{ntk}
	\end{center}
	\vskip -0.2in
\end{figure}

\textbf{Neural Tangent Kernel (NTK)} \label{dpntk}
By substituting the parameter evolution
\begin{eqnarray}
	\frac{\partial\theta^t}{\partial t}=-\frac{\eta}{N}\left[\left.\frac{\partial\mathcal{L}}{\partial f_{\theta}}\right|_{f_{\theta^t},\bm{x}_i}\right]^T_N\cdot \left[\left.\frac{\partial f_{\theta}}{\partial \theta}\right|_{\bm{x}_i,\theta^t}\right]_N
\end{eqnarray}
into the first-order approximation term $(*)$ of Equation~\ref{fparat}, it obtains
\begin{eqnarray}
	(*)&=&\left\langle\left.\frac{\partial f_{\theta}}{\partial \theta}\right|_{\cdot,\theta^t},-\frac{\eta}{N}\left[\left.\frac{\partial\mathcal{L}}{\partial f_{\theta}}\right|_{f_{\theta^t},\bm{x}_i}\right]^T_N\cdot \left[\left.\frac{\partial f_{\theta}}{\partial \theta}\right|_{\bm{x}_i,\theta^t}\right]_N\right\rangle\nonumber\\
	&=&-\frac{\eta}{N}\left[\left.\frac{\partial\mathcal{L}}{\partial f_{\theta}}\right|_{f_{\theta^t},\bm{x}_i}\right]^T_N\cdot\left\langle\left.\frac{\partial f_{\theta}}{\partial \theta}\right|_{\cdot,\theta^t}, \left[\left.\frac{\partial f_{\theta}}{\partial \theta}\right|_{\bm{x}_i,\theta^t}\right]_N\right\rangle\nonumber\\
	&=&-\frac{\eta}{N}\left[\left.\frac{\partial\mathcal{L}}{\partial f_{\theta}}\right|_{f_{\theta^t},\bm{x}_i}\right]^T_N\cdot\left[\left\langle\left.\frac{\partial f_{\theta}}{\partial \theta}\right|_{\cdot,\theta^t},\left.\frac{\partial f_{\theta}}{\partial \theta}\right|_{\bm{x}_i,\theta^t} \right\rangle\right]_N\nonumber\\
	&=&-\frac{\eta}{N}\left[\left.\frac{\partial\mathcal{L}}{\partial f_{\theta}}\right|_{f_{\theta^t},\bm{x}_i}\right]^T_N\cdot\left[K_{\theta^t}(\bm{x}_i,\cdot)\right]_N,
\end{eqnarray}
which derives Equation~\ref{bfparat} as
\begin{eqnarray}
	\frac{\partial f_{\theta^t}}{\partial t}=-\frac{\eta}{N}\left[\left.\frac{\partial\mathcal{L}}{\partial f_{\theta}}\right|_{f_{\theta^t},\bm{x}_i}\right]^T_N\cdot\left[K_{\theta^t}(\bm{x}_i,\cdot)\right]_N+ o\left(\frac{\partial \theta^t}{\partial t}\right),
\end{eqnarray}
and $K_{\theta^t}$ is referred to as neural tangent kernel (NTK)~\cite{jacot2018neural}. Figure~\ref{ntk} provides a visual representation that explains the calculation process of NTK in a clear and understandable way. Informally speaking, studying how a model behaves by focusing on the model itself rather than its parameters typically entails the use of kernel functions.

It can be observed that the quantity $\left.\frac{\partial f_{\theta}}{\partial \theta}\right|_{\cdot,\theta^t}$, present in $K_{\theta^t}(\bm{x}_i,\cdot)=\left\langle\left.\frac{\partial f_{\theta}}{\partial \theta}\right|_{\cdot,\theta^t},\left.\frac{\partial f_{\theta}}{\partial \theta}\right|_{\bm{x}_i,\theta^t} \right\rangle$, represents the partial derivative of the MLP with respect to its parameters, determined by both the structure and specific $\theta^t$, but independent of the input. On the other hand, $\frac{\partial f_{\theta}}{\partial \theta}|_{\bm{x}_i,\theta^t}$originates from the parameter evolution, which relies not only on the MLP structure and specific $\theta^t$, but also on the input example. Assuming the input of $\frac{\partial f_{\theta}}{\partial \theta}|_{\bm{x}_i,\theta^t}$ is not known, the NTK becomes $K_{\theta^t}(\cdot,\cdot)$. On the other hand, if we specify $\bm{x}_j$ as the input for $\frac{\partial f_{\theta}}{\partial \theta}|_{\cdot,\theta^t}$, NTK becomes a scalar as $K_{\theta^t}(\bm{x}_i,\bm{x}_j)=\langle\frac{\partial f_{\theta}}{\partial \theta}|_{\bm{x}_j,\theta^t},\frac{\partial f_{\theta}}{\partial \theta}|_{\bm{x}_i,\theta^t} \rangle$. This indicates that the NTK is a bivariate function represented by $\mathcal{X}\times\mathcal{X}\mapsto\mathbb{R}$, and this form aligns with the kernel used in functional gradient descent. By feeding the input example $\bm{x}_i$, one coordinate of $K_{\theta^t}$ is fixed, causing the MLP to update along $K_{\theta^t}(\bm{x}_i,\cdot)$ based on the magnitude of $\frac{\partial f_{\theta}}{\partial \theta}|_{\bm{x}_i,\theta^t}$
, which is consistent with the underlying mechanism of functional gradient descent. In a nutshell, NTK and the canonical kernel not only maintain consistency in their mathematical representation, but also exhibit alignment in how they influence the evolution of the corresponding MLP. Additionally, Theorem~\ref{ntkconverge} demonstrates the asymptotic relationship between the NTK and the canonical kernel used in functional gradient descent.

\citealp{jacot2018neural} introduce kernel gradient descent, which can be considered as an extension of parameter-based gradient descent. Although kernel gradient descent appears to bear resemblance to functional gradient descent~\cite{zhang2023nonparametric,zhang2023mint}, they fundamentally differ in terms of specific details. In kernel gradient descent, the kernel gradient is derived by incorporating a kernel weighting~\cite{jacot2018neural}, where the NTK serves as the weight to modify the conventional gradient of a real-valued loss $\mathcal{L}(f(\bm{x}),y)$ with respect to $f(\bm{x})$, which is limited to the training set, thus allowing the weighted gradient (kernel gradient) to be extrapolated to values beyond the training set. Differently, functional gradient descent takes a higher-level perspective on the evolution of the MLP in function space~\cite{zhang2023nonparametric,zhang2023mint}. Specifically, $f(\bm{x})=E_{\bm{x}}(f)$ represents the result of evaluating the function $f$ at the example $\bm{x}$, which is defined as the inner product in RKHS between the function $f$ and $K(\bm{x},\cdot)$ (the corresponding kernel with one argument $\bm{x}$) based on the reproducing property. By applying the functional chain rule and Fréchet derivative, the functional gradient is derived accordingly.

Due to the discrete nature of computer operations, functional gradient descent relies on dense pairwise points $\{(\bm{x}_i,K(\bm{x}_\dagger,\bm{x}_i))\}_n$ for representing the kernel $K(\bm{x}_\dagger,\cdot)$, and in order to express $f$, it is necessary to store all $K_{\bm{x}_i}$s as dense points, resulting in significant storage requirements. This issue mirrors the challenge encountered when storing discrete signals, and the solution lies in INR, employing overparameterized MLPs to continuously represent functions, eliminating the need for storing dense points by utilizing a relatively small-sized parameter storage. Besides, in terms of evolution, functional gradient descent requires updating all dense points to derive $f^t$ based on the functional gradient that also relies on $K(\bm{x}_i,\cdot)$, whereas training an MLP only necessitates updating the parameter $\theta$, providing practical convenience compared to the theoretical analysis facilitated by functional gradient descent. This work establishes a correlation between nonparametric teaching and MLP training, which involves training an MLP to represent general functions, thereby increasing the theoretical framework's potential scope for implementation in deep learning.

\begin{figure}[t]
	\vskip 0.2in
	\begin{center}
		\centerline{\includegraphics[width=0.7\columnwidth]{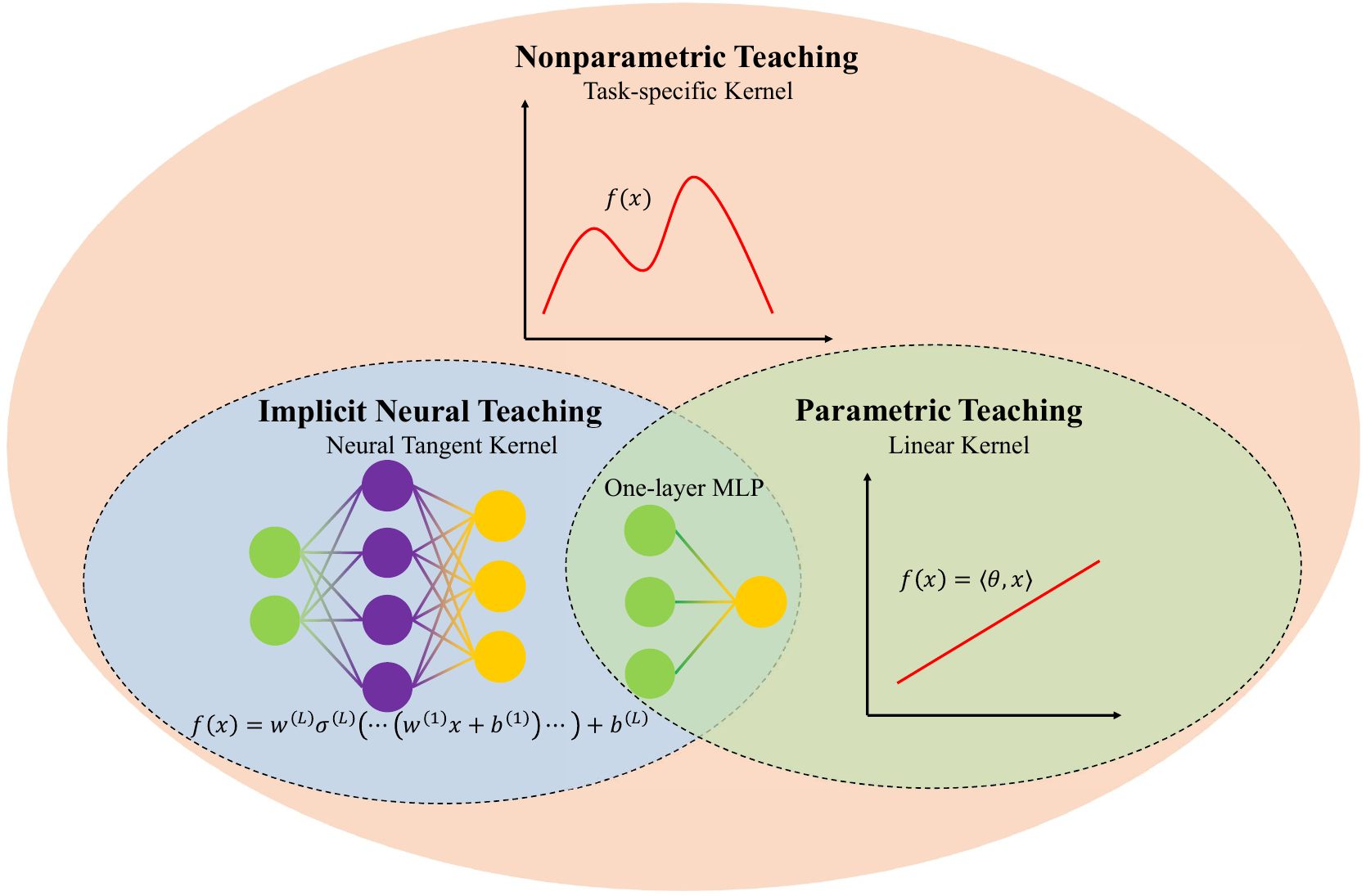}}
		\caption{Illustration of the relationship between nonparametric teaching, implicit neural teaching and parametric teaching. Nonparametric teaching deals with general functions corresponding to task-specific kernels. As an instance, implicit neural teaching focuses on neural tangent kernels~\cite{jacot2018neural} and is concerned with the functions expressed by an overparameterized MLP. On the other hand, parametric teaching concentrates on parameterized functions of the form $f(x)=\langle\theta,\bm{x}\rangle$, which is a specific case of nonparametric teaching that uses a linear kernel as the task-specific kernel. Additionally, teaching a one-layer MLP without nonlinear activation functions is essentially equivalent to parametric teaching.}
		\label{rela}
	\end{center}
	\vskip -0.2in
\end{figure}

\textbf{The Relationship between Nonparametric Teaching, Implicit Neural Teaching, and Parametric Teaching} In simpler terms, nonparametric teaching~\cite{zhang2023nonparametric,ma2019policy} offers a comprehensive framework that encompasses other paradigms, where these paradigms can be viewed as special cases with specific kernels. For instance, this paper focuses on implicit neural teaching, which corresponds to a distinct paradigm by specifying the neural tangent kernel, while parametric teaching~\cite{liu2017iterative,liu2018towards} considers a particular paradigm utilizing a linear kernel. Furthermore, when the MLP is reduced to a single-layer architecture without nonlinear activation functions, it becomes the linear case examined in parametric teaching~\cite{liu2017iterative,liu2018towards}, resulting in a zero remainder in Equation~\ref{fparat}. Figure~\ref{rela} provides a visualization of these relationships.

\textbf{Solution of ODE for training with a fixed single input} If we allow the MLP to evolve based on a single example $\bm{x}$, we have
\begin{eqnarray}
	\frac{\partial f_{\theta^t}}{\partial t} = -\eta(f_{\theta^t}(\bm{x})-f^*(\bm{x}))\cdot K({\bm{x}},\cdot).
\end{eqnarray}
Since $\frac{\partial f^*}{\partial t}=0$, we can rewrite the above differential equation as:
\begin{eqnarray}
	&& \frac{\partial f_{\theta^t}-f^*}{\partial t} = -\eta(f_{\theta^t}(\bm{x})-f^*(\bm{x}))\cdot K({\bm{x}},\cdot).
\end{eqnarray}
By manipulating both sides of the equation using $\left\langle K({\bm{x}},\cdot), \cdot\right\rangle_{\mathcal{H}}$ ($K({\bm{x}},{\bm{x}})\neq0$) and rearranging, we can obtain
\begin{eqnarray}
	&& \mathrm{d}\left(f_{\theta^t}(\bm{x})-f^*(\bm{x})\right) = -\eta(f_{\theta^t}(\bm{x})-f^*(\bm{x}))\cdot K({\bm{x}},{\bm{x}})\mathrm{d}t\nonumber\\
	&&\therefore \frac{\mathrm{d}\left(f_{\theta^t}(\bm{x})-f^*(\bm{x})\right)}{f_{\theta^t}(\bm{x})-f^*(\bm{x})}=-\eta K({\bm{x}},{\bm{x}})\mathrm{d}t\nonumber\\
	&&\therefore \int\frac{\mathrm{d}\left(f_{\theta^t}(\bm{x})-f^*(\bm{x})\right)}{f_{\theta^t}(\bm{x})-f^*(\bm{x})}=-\eta K({\bm{x}},{\bm{x}})\int\mathrm{d}t\nonumber\\
	&&\therefore \ln\left|f_{\theta^t}(\bm{x})-f^*(\bm{x})\right|=-\eta K({\bm{x}},{\bm{x}})t+C.
\end{eqnarray}
When $f_{\theta^t}(\bm{x})$ approaches $f^*(\bm{x})$ from below, that is, $f_{\theta^t}(\bm{x})-f^*(\bm{x})<0$, we have
\begin{eqnarray}
	\ln\left(f^*(\bm{x})-f_{\theta^t}(\bm{x})\right)=-\eta K({\bm{x}},{\bm{x}})t+C.
\end{eqnarray}
Let t=0, we attain
\begin{eqnarray}
	C=\ln\left(f^*(\bm{x})-f_{\theta^0}(\bm{x})\right).
\end{eqnarray}
Therefore, we have
\begin{eqnarray}
	f_{\theta^t}(\bm{x})=f^*(\bm{x})-e^{-\eta K({\bm{x}},{\bm{x}})t}\left(f^*(\bm{x})-f_{\theta^0}(\bm{x})\right).
\end{eqnarray}
If $f_{\theta^t}(\bm{x})$ approaches $f^*(\bm{x})$ from above, which indicates $f_{\theta^t}(\bm{x})-f^*(\bm{x})>0$, we have
\begin{eqnarray}
	f_{\theta^t}(\bm{x})=f^*(\bm{x})+e^{-\eta K({\bm{x}},{\bm{x}})t}\left(f_{\theta^0}(\bm{x})-f^*(\bm{x})\right),
\end{eqnarray}
which is equivalent to the case of $f_{\theta^t}(\bm{x})-f^*(\bm{x})<0$ because
\begin{eqnarray}
	-e^{-\eta K({\bm{x}},{\bm{x}})t}\left(f^*(\bm{x})-f_{\theta^0}(\bm{x})\right)=e^{-\eta K({\bm{x}},{\bm{x}})t}\left(f_{\theta^0}(\bm{x})-f^*(\bm{x})\right).
\end{eqnarray}

\textbf{Detailed solution procedure of matrix ODE corresponding to Equation~\ref{fode}}~\label{dpode} 
The case of $f^*(\bm{x})-f_{\theta^t}(\bm{x})>0$. Since $\frac{\partial f^*}{\partial t}=0$, we can rewrite Equation~\ref{fode}
\begin{eqnarray}
	\frac{\partial f_{\theta^t}}{\partial t} = -\frac{\eta}{N}\left[f_{\theta^t}(\bm{x}_i)-f^*(\bm{x}_i)\right]^T_N\cdot \left[K({\bm{x}_i},\cdot)\right]_N
\end{eqnarray}
as
\begin{eqnarray}
	\frac{\partial f_{\theta^t}-f^*}{\partial t} = -\frac{\eta}{N}\left[f_{\theta^t}(\bm{x}_i)-f^*(\bm{x}_i)\right]^T_N\cdot \left[K({\bm{x}_i},\cdot)\right]_N.
\end{eqnarray}
By applying the inner product $\left\langle\cdot,[K({\bm{x}_j},\cdot)]^T_N\right\rangle_{\mathcal{H}},j\in\mathbb{N}_N$ to both sides of the equation and rearranging, we can derive
\begin{eqnarray}
	&&\mathrm{d}\left(\left[f_{\theta^t}(\bm{x}_j)-f^*(\bm{x}_j)\right]^T_N\right) = -\frac{\eta}{N}\left[f_{\theta^t}(\bm{x}_i)-f^*(\bm{x}_i)\right]^T_N\cdot\left\langle\left[K({\bm{x}_i},\cdot)\right]_N,\left[K({\bm{x}_j},\cdot)\right]^T_N\right\rangle_{\mathcal{H}}\nonumber\\
	&&\therefore \mathrm{d}\left(\left[f_{\theta^t}(\bm{x}_j)-f^*(\bm{x}_j)\right]^T_N\right) = -\frac{\eta}{N}\left[f_{\theta^t}(\bm{x}_i)-f^*(\bm{x}_i)\right]^T_N\cdot \bm{K}\mathrm{d}t,
\end{eqnarray}
where $\bm{K}$ is a symmetric and positive definite matrix of size $N\times N$ with entries $K(\bm{x}_i,\bm{x}_j)$ at the $i$-th row and $j$-th column. By substituting the index $j$ with $i$, we can equivalently derive
\begin{eqnarray}
	\mathrm{d}\left(\left[f_{\theta^t}(\bm{x}_i)-f^*(\bm{x}_i)\right]^T_N\right) = -\frac{\eta}{N}\left[f_{\theta^t}(\bm{x}_i)-f^*(\bm{x}_i)\right]^T_N\cdot \bm{K}\mathrm{d}t,
\end{eqnarray}
which can be expanded version as
\begin{eqnarray}
	&&\mathrm{d}\left[f_{\theta^t}(\bm{x}_1)-f^*(\bm{x}_1),\cdots,f_{\theta^t}(\bm{x}_N)-f^*(\bm{x}_N)\right]\nonumber\\
	&=&-\frac{\eta}{N}\left[f_{\theta^t}(\bm{x}_1)-f^*(\bm{x}_1),\cdots,f_{\theta^t}(\bm{x}_N)-f^*(\bm{x}_N)\right]\nonumber\\
	&&\cdot\left[\begin{array}{cccc}
		K({\bm{x}_1},{\bm{x}_1}) & K({\bm{x}_1},{\bm{x}_2}) & \cdots & K({\bm{x}_1},{\bm{x}_N}) \\
		K({\bm{x}_2},{\bm{x}_1}) & K({\bm{x}_2},{\bm{x}_2}) & \cdots & K({\bm{x}_2},{\bm{x}_N}) \\
		\vdots & \vdots & \ddots & \vdots \\
		K({\bm{x}_N},{\bm{x}_1}) & K({\bm{x}_N},{\bm{x}_2}) & \cdots & K({\bm{x}_N},{\bm{x}_N}) \\
	\end{array}\right]\mathrm{d}t.
\end{eqnarray}
Lemma~\ref{mode} provides the solution for this first-order matrix ordinary differential equation, where $\bm{\alpha}(t)=\left[f_{\theta^t}(\bm{x}_i)-f^*(\bm{x}_i)\right]_N$, $\bm{\alpha}(0)=\left[f_{\theta^0}(\bm{x}_i)-f^*(\bm{x}_i)\right]_N$ and $\bm{A}=\bar{\bm{K}}=\frac{\bm{K}}{N}$, as
\begin{eqnarray}
	\left[f^*(\bm{x}_i)-f_{\theta^t}(\bm{x}_i)\right]^T_N=\left[f^*(\bm{x}_i)-f_{\theta^0}(\bm{x}_i)\right]^T_N\cdot e^{-\eta\bar{\bm{K}}t}.
\end{eqnarray}
We can obtain an equivalent result by transposing it as
\begin{eqnarray}
	\left[f^*(\bm{x}_i)-f_{\theta^t}(\bm{x}_i)\right]_N=e^{-\eta\bar{\bm{K}}t}\cdot\left[f^*(\bm{x}_i)-f_{\theta^0}(\bm{x}_i)\right]_N.
\end{eqnarray}
After rearrangement, it is
\begin{eqnarray}
	\left[f_{\theta^t}(\bm{x}_i)\right]_N=\left[f^*(\bm{x}_i)\right]_N-e^{-\eta\bar{\bm{K}}t}\cdot\left[f^*(\bm{x}_i)-f_{\theta^0}(\bm{x}_i)\right]_N
\end{eqnarray}

For the case of $f^*(\bm{x})-f_{\theta^t}(\bm{x})<0$, similarly, we have
\begin{eqnarray}
	\left[f_{\theta^t}(\bm{x}_i)-f^*(\bm{x}_i)\right]_N=e^{-\eta\bar{\bm{K}}t}\cdot\left[f_{\theta^0}(\bm{x}_i)-f^*(\bm{x}_i)\right]_N.
\end{eqnarray}
After rearrangement, we have
\begin{eqnarray}
	\left[f_{\theta^t}(\bm{x}_i)\right]_N=\left[f^*(\bm{x}_i)\right]_N+e^{-\eta\bar{\bm{K}}t}\cdot\left[f_{\theta^0}(\bm{x}_i)-f^*(\bm{x}_i)\right]_N,
\end{eqnarray}
which is equivalent to the case of $f^*(\bm{x})-f_{\theta^t}(\bm{x})>0$ since
\begin{eqnarray}
	e^{-\eta\bar{\bm{K}}t}\cdot\left[f_{\theta^0}(\bm{x}_i)-f^*(\bm{x}_i)\right]_N=-e^{-\eta\bar{\bm{K}}t}\cdot\left[f^*(\bm{x}_i)-f_{\theta^0}(\bm{x}_i)\right]_N.
\end{eqnarray}
This concludes the solution.

In the sense that a function can be seen as an infinite-dimensional generalization of a Euclidean vector, Equation~\ref{smode} can be generalized as:
\begin{eqnarray*}
		f_{\theta^t}(\cdot)=f^*(\cdot)+\sum_{i=1}^{\infty}e^{-\eta\lambda_i t}\nu_i(\cdot)\cdot\underbrace{\left\langle\nu_i,(f_{\theta^0}-f^*)\right\rangle_{\mathcal{H}}}_{\text{It is a constant}},
\end{eqnarray*}
where $\nu_i$ denotes the corresponding eigenfunction based on spectral decomposition, \ie, infinite eigenvectors.

\clearpage
\section{Detailed Proofs}
\textbf{Proof of Theorem~\ref{ntkconverge}}\label{pntkconverge} By representing the evolution of an MLP through the variation of parameters and through a high-level standpoint of function variation, we have
\begin{eqnarray}
	-\frac{\eta}{N}\left[\left.\frac{\partial\mathcal{L}}{\partial f_{\theta}}\right|_{f_{\theta^t},\bm{x}_i}\right]^T_N\cdot \left[K({\bm{x}_i},\cdot)\right]_N=
	-\frac{\eta}{N}\left[\left.\frac{\partial\mathcal{L}}{\partial f_{\theta}}\right|_{f_{\theta^t},\bm{x}_i}\right]^T_N\cdot\left[\left\langle\left.\frac{\partial f_{\theta}}{\partial \theta}\right|_{\cdot,\theta^t},\left.\frac{\partial f_{\theta}}{\partial \theta}\right|_{\bm{x}_i,\theta^t} \right\rangle\right]_N + o\left(\frac{\partial \theta^t}{\partial t}\right).
\end{eqnarray}
Following the reorganization, we obtain
\begin{eqnarray}
	-\frac{\eta}{N}\left[\left.\frac{\partial\mathcal{L}}{\partial f_{\theta}}\right|_{f_{\theta^t},\bm{x}_i}\right]^T_N\cdot \left[K({\bm{x}_i},\cdot)-K_{\theta^t}(\bm{x}_i,\cdot)\right]_N=o\left(\frac{\partial \theta^t}{\partial t}\right).
\end{eqnarray}
By substituting the evolution of the parameters
\begin{eqnarray}
	\frac{\partial\theta^t}{\partial t}=-\eta\frac{\partial\mathcal{L}}{\partial\theta^t}=-\frac{\eta}{N}\left[\left.\frac{\partial\mathcal{L}}{\partial f_{\theta}}\right|_{f_{\theta^t},\bm{x}_i}\right]^T_N\cdot \left[\left.\frac{\partial f_{\theta}}{\partial \theta}\right|_{\bm{x}_i,\theta^t}\right]_N
\end{eqnarray}
into the remainder, we obtain
\begin{eqnarray}
	-\frac{\eta}{N}\left[\left.\frac{\partial\mathcal{L}}{\partial f_{\theta}}\right|_{f_{\theta^t},\bm{x}_i}\right]^T_N\cdot \left[K({\bm{x}_i},\cdot)-K_{\theta^t}(\bm{x}_i,\cdot)\right]_N=o\left(-\frac{\eta}{N}\left[\left.\frac{\partial\mathcal{L}}{\partial f_{\theta}}\right|_{f_{\theta^t},\bm{x}_i}\right]^T_N\cdot \left[\left.\frac{\partial f_{\theta}}{\partial \theta}\right|_{\bm{x}_i,\theta^t}\right]_N\right).
\end{eqnarray}
During the training of an MLP with a convex loss $\mathcal{L}$ (which is convex with respect to $f_\theta$ but usually nonconvex with respect to $\theta$), we have the limit of the vector $\lim_{t\to\infty}\left[\left.\frac{\partial\mathcal{L}}{\partial f_{\theta}}\right|_{f_{\theta^t},\bm{x}_i}\right]_N=\bm{0}$. Since the right-hand side of the equation is of a higher order infinitesimal compared to the left-hand side, to maintain this equality, we can conclude that
\begin{eqnarray}
	\lim_{t\to\infty}\left[K({\bm{x}_i},\cdot)-K_{\theta^t}(\bm{x}_i,\cdot)\right]_N=\bm{0}.
\end{eqnarray}
This implies that for each $\bm{x}\in\{\bm{x}_i\}_N$, NTK converges point-wise to the canonical kernel.

$\blacksquare$

\textbf{Proof of Proposition~\ref{slr}} \label{pslr}
By recollecting the definition of Fréchet derivative in Definition~\ref{fd}, the convexity of $\mathcal{L}$ implies that
\begin{eqnarray}
	\frac{\partial\mathcal{L}}{\partial t}\leq\underbrace{\left\langle\frac{\partial\mathcal{L}}{\partial f_{\theta^{t+1}}},\frac{f_{\theta^t}}{\partial t}\right\rangle_\mathcal{H}}_{\Xi}.
\end{eqnarray}
By specifying the Fréchet derivative of $\frac{\partial\mathcal{L}}{\partial f_{\theta^{t+1}}}$ and the evolution of $f_{\theta^t}$, the r.h.s. term $\Xi$ can be expressed as
\begin{eqnarray}\label{eqxi}
	\Xi&=&\left\langle\mathcal{G}^{t+1},-\eta\mathcal{G}^t\right\rangle_{\mathcal{H}}\nonumber\\
	&=&\frac{\eta}{N^2}\left\langle\left[\left.\frac{\partial\mathcal{L}}{\partial f_{\theta}}\right|_{f_{\theta^{t+1}},\bm{x}_i}\right]^T_N\cdot \left[K_{\bm{x}_i}\right]_N,[K_{\bm{x}_i}]^T_N\cdot\left[\left.\frac{\partial\mathcal{L}}{\partial f_{\theta}}\right|_{f_{\theta^t},\bm{x}_i}\right]_N\right\rangle_{\mathcal{H}}\nonumber\\
	&=&-\frac{\eta}{N^2}\left[\left.\frac{\partial\mathcal{L}}{\partial f_{\theta}}\right|_{f_{\theta^{t+1}},\bm{x}_i}\right]^T_N\cdot \left\langle\left[K_{\bm{x}_i}\right]_N,[K_{\bm{x}_i}]^T_N\right\rangle_{\mathcal{H}}\cdot\left[\left.\frac{\partial\mathcal{L}}{\partial f_{\theta}}\right|_{f_{\theta^t},\bm{x}_i}\right]_N\nonumber\\
	&=&-\frac{\eta}{N}\left[\left.\frac{\partial\mathcal{L}}{\partial f_{\theta}}\right|_{f_{\theta^t},\bm{x}_i}\right]^T_N\bar{\bm{K}}\left[\left.\frac{\partial\mathcal{L}}{\partial f_{\theta}}\right|_{f_{\theta^{t+1}},\bm{x}_i}\right]_N,
\end{eqnarray}
where $\bar{\bm{K}}=\bm{K}/N$, and $\bm{K}$ is a symmetric and positive definite matrix of size $N\times N$ with elements $K(\bm{x}_i,\bm{x}_j)$ located at the $i$-th row and $j$-th column. Furthermore, the last term in Equation~\ref{eqxi} can be rewritten as
\begin{eqnarray}\label{epxi}
	&&-\frac{\eta}{N}\left[\left.\frac{\partial\mathcal{L}}{\partial f_{\theta}}\right|_{f_{\theta^t},\bm{x}_i}\right]^T_N\bar{\bm{K}}\left[\left.\frac{\partial\mathcal{L}}{\partial f_{\theta}}\right|_{f_{\theta^{t+1}},\bm{x}_i}\right]_N\nonumber\\
	&=&-\frac{\eta}{N}\left[\left.\frac{\partial\mathcal{L}}{\partial f_{\theta}}\right|_{f_{\theta^t},\bm{x}_i}\right]^T_N\bar{\bm{K}}\left(\left[\left.\frac{\partial\mathcal{L}}{\partial f_{\theta}}\right|_{f_{\theta^{t+1}},\bm{x}_i}\right]_N+\left[\left.\frac{\partial\mathcal{L}}{\partial f_{\theta}}\right|_{f_{\theta^t},\bm{x}_i}\right]_N-\left[\left.\frac{\partial\mathcal{L}}{\partial f_{\theta}}\right|_{f_{\theta^t},\bm{x}_i}\right]_N\right)\nonumber\\
	&=&-\frac{\eta}{N}\left[\left.\frac{\partial\mathcal{L}}{\partial f_{\theta}}\right|_{f_{\theta^t},\bm{x}_i}\right]^T_N\bar{\bm{K}}\left[\left.\frac{\partial\mathcal{L}}{\partial f_{\theta}}\right|_{f_{\theta^t},\bm{x}_i}\right]_N-\frac{\eta}{N}\left[\left.\frac{\partial\mathcal{L}}{\partial f_{\theta}}\right|_{f_{\theta^t},\bm{x}_i}\right]^T_N\bar{\bm{K}}\left(\left[\left.\frac{\partial\mathcal{L}}{\partial f_{\theta}}\right|_{f_{\theta^{t+1}},\bm{x}_i}\right]_N-\left[\left.\frac{\partial\mathcal{L}}{\partial f_{\theta}}\right|_{f_{\theta^t},\bm{x}_i}\right]_N\right)\nonumber\\
	&=&-\frac{\eta}{N}\left[\left.\frac{\partial\mathcal{L}}{\partial f_{\theta}}\right|_{f_{\theta^t},\bm{x}_i}\right]^T_N\bar{\bm{K}}\left[\left.\frac{\partial\mathcal{L}}{\partial f_{\theta}}\right|_{f_{\theta^t},\bm{x}_i}\right]_N\nonumber\\
	&&+\frac{\eta}{N}\left(\left[\left.\frac{\partial\mathcal{L}}{\partial f_{\theta}}\right|_{f_{\theta^{t+1}},\bm{x}_i}\right]^T_N-\left[\left.\frac{\partial\mathcal{L}}{\partial f_{\theta}}\right|_{f_{\theta^t},\bm{x}_i}\right]^T_N-\left[\left.\frac{\partial\mathcal{L}}{\partial f_{\theta}}\right|_{f_{\theta^{t+1}},\bm{x}_i}\right]^T_N\right)\bar{\bm{K}}\left(\left[\left.\frac{\partial\mathcal{L}}{\partial f_{\theta}}\right|_{f_{\theta^{t+1}},\bm{x}_i}\right]_N-\left[\left.\frac{\partial\mathcal{L}}{\partial f_{\theta}}\right|_{f_{\theta^t},\bm{x}_i}\right]_N\right).
\end{eqnarray}
The last term in Equation~\ref{epxi} above can be elaborated as
\begin{eqnarray}\label{lstepxi}
	&&\frac{\eta}{N}\left(\left[\left.\frac{\partial\mathcal{L}}{\partial f_{\theta}}\right|_{f_{\theta^{t+1}},\bm{x}_i}\right]^T_N-\left[\left.\frac{\partial\mathcal{L}}{\partial f_{\theta}}\right|_{f_{\theta^t},\bm{x}_i}\right]^T_N-\left[\left.\frac{\partial\mathcal{L}}{\partial f_{\theta}}\right|_{f_{\theta^{t+1}},\bm{x}_i}\right]^T_N\right)\bar{\bm{K}}\left(\left[\left.\frac{\partial\mathcal{L}}{\partial f_{\theta}}\right|_{f_{\theta^{t+1}},\bm{x}_i}\right]_N-\left[\left.\frac{\partial\mathcal{L}}{\partial f_{\theta}}\right|_{f_{\theta^t},\bm{x}_i}\right]_N\right)\nonumber\\
	&=&\frac{\eta}{N}\left(\left[\left.\frac{\partial\mathcal{L}}{\partial f_{\theta}}\right|_{f_{\theta^{t+1}},\bm{x}_i}\right]_N-\left[\left.\frac{\partial\mathcal{L}}{\partial f_{\theta}}\right|_{f_{\theta^t},\bm{x}_i}\right]_N\right)^T\bar{\bm{K}}\left(\left[\left.\frac{\partial\mathcal{L}}{\partial f_{\theta}}\right|_{f_{\theta^{t+1}},\bm{x}_i}\right]_N-\left[\left.\frac{\partial\mathcal{L}}{\partial f_{\theta}}\right|_{f_{\theta^t},\bm{x}_i}\right]_N\right)\nonumber\\
	&&-\frac{\eta}{N}\left[\left.\frac{\partial\mathcal{L}}{\partial f_{\theta}}\right|_{f_{\theta^{t+1}},\bm{x}_i}\right]^T_N\bar{\bm{K}}\left(\left[\left.\frac{\partial\mathcal{L}}{\partial f_{\theta}}\right|_{f_{\theta^{t+1}},\bm{x}_i}\right]_N-\left[\left.\frac{\partial\mathcal{L}}{\partial f_{\theta}}\right|_{f_{\theta^t},\bm{x}_i}\right]_N\right)\nonumber\\
	&=&\frac{\eta}{N}\left[\left.\frac{\partial\mathcal{L}}{\partial f_{\theta}}\right|_{f_{\theta^{t+1}},\bm{x}_i}-\left.\frac{\partial\mathcal{L}}{\partial f_{\theta}}\right|_{f_{\theta^t},\bm{x}_i}\right]_N^T\bar{\bm{K}}\left[\left.\frac{\partial\mathcal{L}}{\partial f_{\theta}}\right|_{f_{\theta^{t+1}},\bm{x}_i}-\left.\frac{\partial\mathcal{L}}{\partial f_{\theta}}\right|_{f_{\theta^t},\bm{x}_i}\right]_N\nonumber\\
	&&-\frac{\eta}{N}\left(\left[\left.\frac{\partial\mathcal{L}}{\partial f_{\theta}}\right|_{f_{\theta^{t+1}},\bm{x}_i}\right]_N-\frac{1}{2}\left[\left.\frac{\partial\mathcal{L}}{\partial f_{\theta}}\right|_{f_{\theta^t},\bm{x}_i}\right]_N\right)^T\bar{\bm{K}}\left(\left[\left.\frac{\partial\mathcal{L}}{\partial f_{\theta}}\right|_{f_{\theta^{t+1}},\bm{x}_i}\right]_N-\frac{1}{2}\left[\left.\frac{\partial\mathcal{L}}{\partial f_{\theta}}\right|_{f_{\theta^t},\bm{x}_i}\right]_N\right)\nonumber\\
	&&+\frac{\eta}{4N}\left[\left.\frac{\partial\mathcal{L}}{\partial f_{\theta}}\right|_{f_{\theta^t},\bm{x}_i}\right]^T_N\bar{\bm{K}}\left[\left.\frac{\partial\mathcal{L}}{\partial f_{\theta}}\right|_{f_{\theta^t},\bm{x}_i}\right]_N.
\end{eqnarray}
Since $\bar{\bm{K}}$ is positive definite, it is clear that $\frac{\eta}{N}\left(\left[\left.\frac{\partial\mathcal{L}}{\partial f_{\theta}}\right|_{f_{\theta^{t+1}},\bm{x}_i}\right]_N-\frac{1}{2}\left[\left.\frac{\partial\mathcal{L}}{\partial f_{\theta}}\right|_{f_{\theta^t},\bm{x}_i}\right]_N\right)^T\bar{\bm{K}}\left(\left[\left.\frac{\partial\mathcal{L}}{\partial f_{\theta}}\right|_{f_{\theta^{t+1}},\bm{x}_i}\right]_N-\frac{1}{2}\left[\left.\frac{\partial\mathcal{L}}{\partial f_{\theta}}\right|_{f_{\theta^t},\bm{x}_i}\right]_N\right)$ is a non-negative term, and therefore by combining Equation~\ref{eqxi}, \ref{epxi}, and \ref{lstepxi}, we have
\begin{eqnarray}\label{xileq}
	\Xi&\leq&-\frac{3\eta}{4N}\underbrace{\left[\left.\frac{\partial\mathcal{L}}{\partial f_{\theta}}\right|_{f_{\theta^t},\bm{x}_i}\right]^T_N\bar{\bm{K}}\left[\left.\frac{\partial\mathcal{L}}{\partial f_{\theta}}\right|_{f_{\theta^t},\bm{x}_i}\right]_N}_{\textcircled{\scriptsize 1}}\nonumber\\
	&&+\frac{\eta}{N}\underbrace{\left[\left.\frac{\partial\mathcal{L}}{\partial f_{\theta}}\right|_{f_{\theta^{t+1}},\bm{x}_i}-\left.\frac{\partial\mathcal{L}}{\partial f_{\theta}}\right|_{f_{\theta^t},\bm{x}_i}\right]_N^T\bar{\bm{K}}\left[\left.\frac{\partial\mathcal{L}}{\partial f_{\theta}}\right|_{f_{\theta^{t+1}},\bm{x}_i}-\left.\frac{\partial\mathcal{L}}{\partial f_{\theta}}\right|_{f_{\theta^t},\bm{x}_i}\right]_N}_{\textcircled{\scriptsize2}}.
\end{eqnarray}

Given the evaluation functional definition and the assumption that $\mathcal{L}$ is Lipschitz smooth with a constant $\xi>0$, the term $\textcircled{\scriptsize2}$ in the last term of Equation~\ref{xileq} is upper bounded as
\begin{eqnarray}\label{ls}
	\textcircled{\scriptsize2}&=&\left[\left.\frac{\partial\mathcal{L}}{\partial f_{\theta}}\right|_{f_{\theta^{t+1}},\bm{x}_i}-\left.\frac{\partial\mathcal{L}}{\partial f_{\theta}}\right|_{f_{\theta^t},\bm{x}_i}\right]_N^T\bar{\bm{K}}\left[\left.\frac{\partial\mathcal{L}}{\partial f_{\theta}}\right|_{f_{\theta^{t+1}},\bm{x}_i}-\left.\frac{\partial\mathcal{L}}{\partial f_{\theta}}\right|_{f_{\theta^t},\bm{x}_i}\right]_N\nonumber\\
	&=&\left[E_{\bm{x}_i}\left(\left.\frac{\partial\mathcal{L}}{\partial f_{\theta}}\right|_{f_{\theta^{t+1}}}-\left.\frac{\partial\mathcal{L}}{\partial f_{\theta}}\right|_{f_{\theta^t}}\right)\right]^T_N\bar{\bm{K}}\left[E_{\bm{x}_i}\left(\left.\frac{\partial\mathcal{L}}{\partial f_{\theta}}\right|_{f_{\theta^{t+1}}}-\left.\frac{\partial\mathcal{L}}{\partial f_{\theta}}\right|_{f_{\theta^t}}\right)\right]_N\nonumber\\
	&\leq&\xi^2\left[E_{\bm{x}_i}\left(f_{\theta^{t+1}}-f_{\theta^t}\right)\right]^T_N\bar{\bm{K}}\left[E_{\bm{x}_i}\left(f_{\theta^{t+1}}-f_{\theta^t}\right)\right]_N\nonumber\\
	&=&\xi^2\left\langle\left(f_{\theta^{t+1}}-f_{\theta^t}\right),\left[K_{\bm{x}_i}\right]^T_N\right\rangle_{\mathcal{H}}\cdot\bar{\bm{K}}\cdot\left\langle\left[K_{\bm{x}_i}\right]_N,\left(f_{\theta^{t+1}}-f_{\theta^t}\right)\right\rangle_{\mathcal{H}}\nonumber\\
	&=&{\eta}^2\xi^2\cdot\left[\left.\frac{\partial\mathcal{L}}{\partial f_{\theta}}\right|_{f_{\theta^t},\bm{x}_i}\right]_N^T\frac{\left\langle\left[K_{\bm{x}_i}\right]_N,[K_{\bm{x}_i}]^T_N\right\rangle_{\mathcal{H}}}{N}\cdot\bar{\bm{K}}\cdot\frac{\left\langle\left[K_{\bm{x}_i}\right]_N,[K_{\bm{x}_i}]^T_N\right\rangle_{\mathcal{H}}}{N}\cdot\left[\left.\frac{\partial\mathcal{L}}{\partial f_{\theta}}\right|_{f_{\theta^t},\bm{x}_i}\right]_N.
\end{eqnarray}
Based on the assumption that the canonical kernel is bounded above by a constant $\zeta>0$, we have 
\begin{eqnarray}
	\left\langle\left[K_{\bm{x}_i}\right]_N,[K_{\bm{x}_i}]^T_N\right\rangle_{\mathcal{H}}\leq\zeta\left\langle [1]_N,[1]^T_N\right\rangle,\nonumber
\end{eqnarray}
and 
\begin{eqnarray}
	\bar{\bm{K}}\leq\frac{\zeta}{N}\left\langle [1]_N,[1]^T_N\right\rangle. \nonumber
\end{eqnarray} 
Therefore, $\textcircled{\scriptsize1}$ is bounded above as
\begin{eqnarray}\label{bk1}
	\textcircled{\scriptsize1}&\leq&\frac{\zeta}{N}\left\langle\left[\sum_{i=1}^N\left.\frac{\partial\mathcal{L}}{\partial f_{\theta}}\right|_{f_{\theta^t},\bm{x}_i}\right]^T_N,[1]_N\right\rangle\left\langle[1]_N^T,\left[\sum_{i=1}^N\left.\frac{\partial\mathcal{L}}{\partial f_{\theta}}\right|_{f_{\theta^t},\bm{x}_i}\right]_N\right\rangle\nonumber\\
	&=&\frac{\zeta}{N}\left(\sum_{i=1}^N\left.\frac{\partial\mathcal{L}}{\partial f_{\theta}}\right|_{f_{\theta^t},\bm{x}_i}\right)^2.
\end{eqnarray}
Simultaneously, the last term in Equation~\ref{ls} is also bounded from above:
\begin{eqnarray}\label{bk2}
	&&{\eta}^2\xi^2\cdot\left[\left.\frac{\partial\mathcal{L}}{\partial f_{\theta}}\right|_{f_{\theta^t},\bm{x}_i}\right]_N^T\frac{\left\langle\left[K_{\bm{x}_i}\right]_N,[K_{\bm{x}_i}]^T_N\right\rangle_{\mathcal{H}}}{N}\cdot\bar{\bm{K}}\cdot\frac{\left\langle\left[K_{\bm{x}_i}\right]_N,[K_{\bm{x}_i}]^T_N\right\rangle_{\mathcal{H}}}{N}\cdot\left[\left.\frac{\partial\mathcal{L}}{\partial f_{\theta}}\right|_{f_{\theta^t},\bm{x}_i}\right]_N\nonumber\\
	&\leq&{\eta}^2\xi^2\left[\frac{\zeta}{N}\sum_{i=1}^N\left.\frac{\partial\mathcal{L}}{\partial f_{\theta}}\right|_{f_{\theta^t},\bm{x}_i}\right]^T\cdot\bar{\bm{K}}\cdot\left[\frac{\zeta}{N}\sum_{i=1}^N\left.\frac{\partial\mathcal{L}}{\partial f_{\theta}}\right|_{f_{\theta^t},\bm{x}_i}\right]_N\nonumber\\
	&\leq&\frac{{\eta}^2\xi^2\zeta^3}{N}\left\langle\left[\frac{1}{N}\sum_{i=1}^N\left.\frac{\partial\mathcal{L}}{\partial f_{\theta}}\right|_{f_{\theta^t},\bm{x}_i}\right]^T_N,[1]_N\right\rangle\left\langle[1]_N^T,\left[\frac{1}{N}\sum_{i=1}^N\left.\frac{\partial\mathcal{L}}{\partial f_{\theta}}\right|_{f_{\theta^t},\bm{x}_i}\right]_N\right\rangle\nonumber\\
	&=&\frac{{\eta}^2\xi^2\zeta^3}{N}\left(\sum_{i=1}^N\left.\frac{\partial\mathcal{L}}{\partial f_{\theta}}\right|_{f_{\theta^t},\bm{x}_i}\right)^2.
\end{eqnarray}
Therefore, by combining Equations~\ref{xileq}, \ref{ls}, \ref{bk1}, and \ref{bk2}, we obtain
\begin{eqnarray}
	\Xi\leq-\eta\zeta\left(\frac{3}{4}-{\eta}^2\xi^2\zeta^2\right)\left(\frac{1}{N}\sum_{i=1}^N\left.\frac{\partial\mathcal{L}}{\partial f_{\theta}}\right|_{f_{\theta^t},\bm{x}_i}\right)^2,
\end{eqnarray}
which indicates
\begin{eqnarray}
	\frac{\partial \mathcal{L}}{\partial t}\leq\Xi\leq-\eta\zeta\left(\frac{3}{4}-{\eta}^2\xi^2\zeta^2\right)\left(\frac{1}{N}\sum_{i=1}^N\left.\frac{\partial\mathcal{L}}{\partial f_{\theta}}\right|_{f_{\theta^t},\bm{x}_i}\right)^2.
\end{eqnarray}
Hence, if $\eta\leq\frac{1}{2\xi\zeta}$, we have
\begin{eqnarray}
	\frac{\partial \mathcal{L}}{\partial t}\leq -\frac{\eta\zeta}{2}\left(\frac{1}{N}\sum_{i=1}^N\left.\frac{\partial\mathcal{L}}{\partial f_{\theta}}\right|_{f_{\theta^t},\bm{x}_i}\right)^2.
\end{eqnarray}

$\blacksquare$

\textbf{Proof of Lemma~\ref{mode}} \label{pmode} For $\bm{\alpha}(t)=e^{\bm{A}t}\bm{c}$, where $e^{\bm{A}t}=\sum_{i=0}^{\infty}\frac{t^i\bm{A}^i}{i!}$ and $\bm{c}$ is a time-independent column vector of size $n\times1$, we have 
\begin{eqnarray}
	\frac{\partial\bm{\alpha}(t)}{\partial t}&=&\frac{\partial e^{\bm{A}t}\bm{c}}{\partial t}=\frac{\partial \sum_{i=0}^{\infty}\frac{t^i\bm{A}^i}{i!}\bm{c}}{\partial t}=\sum_{i=1}^{\infty}\frac{\partial t^i}{\partial t}\frac{\bm{A}^i\bm{c}}{i!}\nonumber\\
	&=&\bm{A}\sum_{i=1}^{\infty}\frac{\bm{A}^{i-1}t^{i-1}\bm{c}}{(i-1)!}=\bm{A}\sum_{i=0}^{\infty}\frac{\bm{A}^{i}t^{i}\bm{c}}{i!}=\bm{A}e^{\bm{A}t}\bm{c}=\bm{A}\bm{\alpha}(t).
\end{eqnarray}
Meanwhile, by setting $t=0$, we have 
\begin{eqnarray}
	\bm{\alpha}(0)=e^{0}\bm{c},
\end{eqnarray}
which means $\bm{c}=\bm{\alpha}(0)$. Therefore, $\bm{\alpha}(t)=e^{\bm{A}t}\bm{\alpha}(0)$ is the unique solution of the matrix ODE $\frac{\partial\bm{\alpha}(t)}{\partial t}=\bm{A}\bm{\alpha}(t)$ with initial value $\bm{\alpha}(0)$.

$\blacksquare$

\clearpage
\section{Experiment Details}
\label{ade}
\subsection{Synthetic 1D signal}
The FFN consists of 4 layers, each with 256 hidden units, and the value of $\sigma$ is set to 2 for the random Fourier features used in the FFN. Based on Theorem~\ref{ntkconverge}, the canonical kernel used in FGD is approximated by adopting the empirical NTK of the INR obtained through PGD after 5000 iterations.

\subsection{Toy 2D Cameraman Fitting}
We train SIREN models with 6 layers, each with 256 hidden units, with default settings as mentioned in \citealp{sitzmann2020implicit} to fit the 512$\times$512 Cameraman grayscale image from \texttt{scikit-image}~\cite{van2014scikit}. To have a close resemblance with the theoretical analysis of INT, we train the models with vanilla gradient descent without momentum for 5000 iterations. All models are trained using a cosine annealing scheduler with a starting learning rate of 1e-4 and a minimum learning rate of 1e-6. The specific INT sampling strategies of the 4 different SIREN models presented in Figure \ref{fig:sgd_pred} are as follows:
\begin{itemize}
    \item w/o INT - At each optimization step, the entire image is used.
    \item w/o INT (20\%) - At each optimization step, a random 20\% of pixels are used.
    \item With INT (20\%) - At each optimization step, pixels with the top 20\% error rates from the previous training iteration are used to train the current iteration.
    \item With INT (incre.) - Similar scheme as ``with INT (20\%)'', except that we increase the sampling rate by 8\% for every 500 iterations from 20\% to 92\%.
\end{itemize}

\subsection{INT Strategy Experiment}
\begin{figure}[h!]
    \centering
    \includegraphics[width=\linewidth]{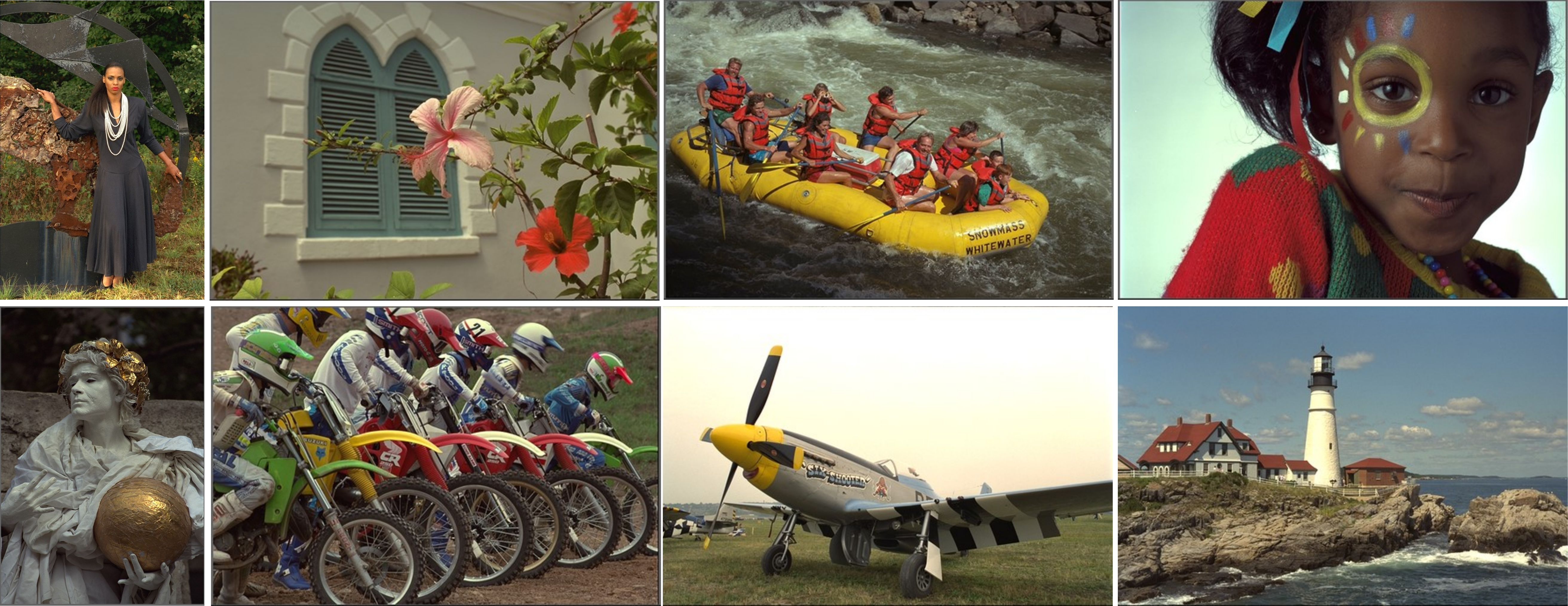}
    \caption{The selected 8/24 images from the Kodak dataset.}
    \label{fig:kodak_subset}
\end{figure}

We train identical SIREN models as mentioned in the previous section on 8/24 images from the Kodak dataset~\cite{kodak}. As numerous strategies were tested and we hoped to utilize a wide variety of images to find a robust strategy that works not only across different image datasets but also other modalities, we chose only a representative subset of the Kodak dataset for experimental efficiency. As shown in Figure \ref{fig:kodak_subset}, this subset of images is chosen to include both simple images (e.g. single object), complex images (e.g. multiple objects or high-frequency signals such as grass), and images with humans. To better simulate real-world scenarios of utilizing INRs, We test our strategies with the Adam \cite{kingma2014adam} optimizer with a learning rate of 1e-3 and an identical cosine annealing scheduler for the learning rate as in the previous section. All models are trained for 5000 iterations.

We highlight that logging PSNR/SSIM values and saving visualization results during training takes up significant time. Thus, to record the most realistic training time, we retrain all the models with the same seed and configurations but without any logs except for the loss value on a single image. As all images have the same dimensions, this is sufficient to represent the general trend of training times across the strategies.

The specific INT strategies presented in Figure \ref{fig:strategy} are as follows:
\begin{itemize}
    \item Ratio
    \begin{itemize}
        \item Cosine - Increasing sampling ratio from 20\% to 100\% in a cosine annealing manner.
        \item R-Cosine -  Decreasing sampling ratio from 100\% to 20\% in a cosine annealing manner.
        \item Step - Incrementing sampling ratio from 20\% to 92\% in 10 equal intervals, which is 500 iterations in this case where we train for a total of 5000 iterations.
    \end{itemize}
    \item Interval
    \begin{itemize}
        \item Dense - Sample points with top \textless ratio\textgreater\% error rates for \textit{every training iteration}. Note that the error rates are obtained from the previous iteration.
        \item Decremental - Sampling interval decreases from every 90 iterations to 1 iteration incrementally in 10 intervals, which is 500 iterations in this case where we train for a total of 5000 iterations. That is, at every 500 iterations, we decrease the interval by 10, except for the last 500 iterations where we decrease by 9 from 10 to 1.
        \item Incremental - Sampling interval increases from every 1 iteration to 90 iterations incrementally in 10 intervals, which is 500 iterations in this case where we train for a total of 5000 iterations. That is, at every 500 iterations, we increase the interval by 10, except for the first 500 iterations where we increase by 9 from 1 to 10.
    \end{itemize}
\end{itemize}

\begin{figure}[h!]
    \centering
    \includegraphics[width=\linewidth]{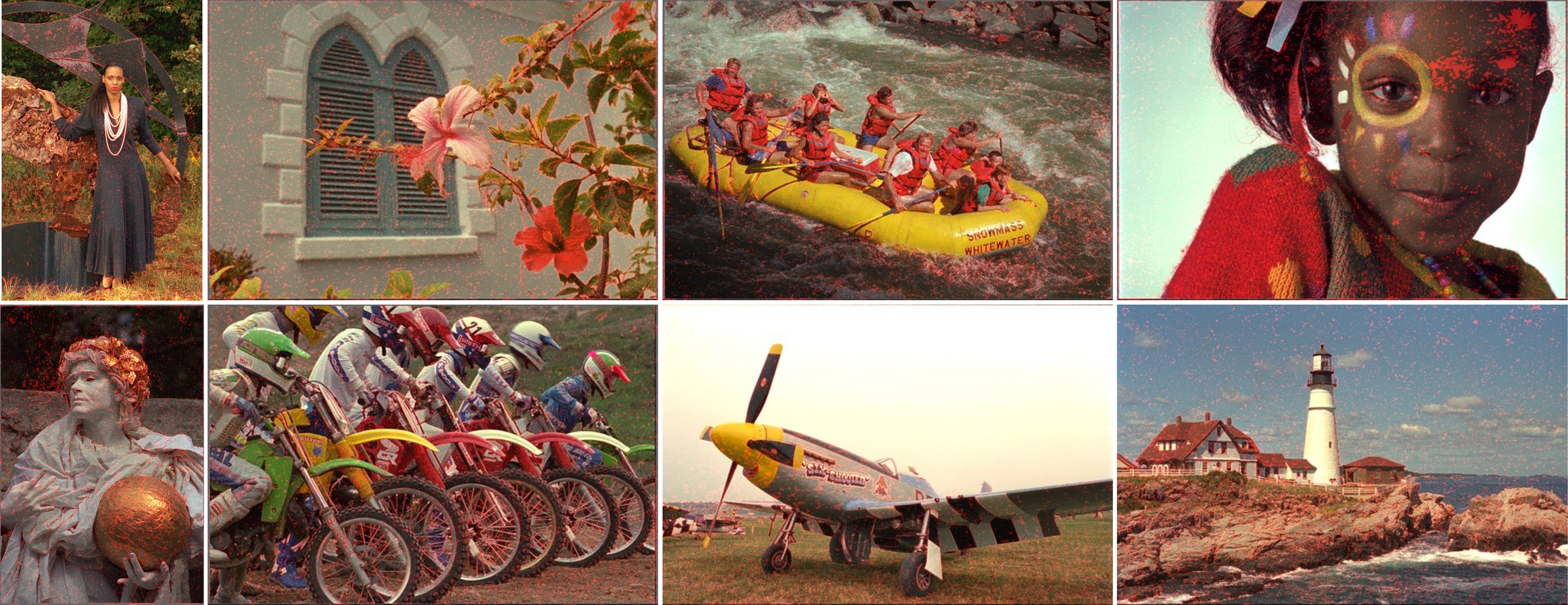}
    \caption{The selected 8/24 images from the Kodak dataset and its selected training points at a particular instance.}
    \label{fig:kodak_subset_selected}
\end{figure}

\begin{figure}[h!]
    \centering
    \includegraphics[width=\linewidth]{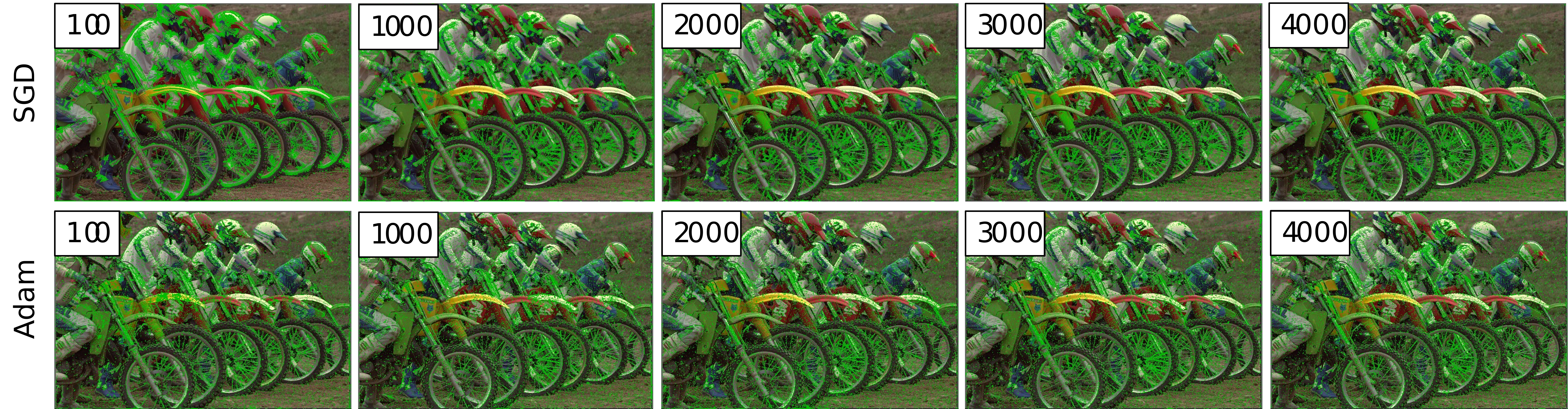}
    \caption{Visualizing the progression of sampled points when trained with SGD vs Adam on kodak05 image.}
    \label{fig:sgd_vs_adam_sampling}
\end{figure}

Figure \ref{fig:sgd_vs_adam_sampling} presents the sampling progression of SIREN trained with SGD and Adam on the kodak05 image. Besides, examining the applicability of active data selection methods~\cite{loshchilov2015online, graves2017automated, mindermann2022prioritized} for INRs learning efficiency could be interesting.

\subsection{Multi-modality Signal Fitting}

For all modalities, we train a SIREN model with Adam optimizer and cosine annealing learning rate scheduler. We set $\omega_0=30$ for the SIREN model. All modalities except 2D Kodak images start with a learning rate of 1e-4, while 2D Kodak images start with 1e-3. We select the best INT strategy found in the previous section to train for all modalities: ``step-incremental''. Note that we always partition the training into 10 same-sized intervals where each interval has its respective INT sampling ratio and sampling interval. For instance, if we train audio samples for 10K iterations, then we start with 20\% sampling ratio and a sampling ratio of 1 and progressively add 8\% to the sampling ratio and 10 to the sampling interval for every 1K iterations.

\textbf{1D Audio.} The Librispeech dataset~\cite{panayotov2015librispeech} is chosen for the audio-fitting task. We select the first 100 samples from the test-clean split that have a duration greater than 2 seconds. For our evaluation benchmark, we clip the first 2 seconds of each sample. We train a SIREN with 5 layers, each having 128 hidden units, resulting in a total of approximately 50K parameters. Each sample is trained for 10K iterations.

\textbf{2D image.} The entire Kodak dataset~\cite{kodak} is used in this case. Model configuration and training parameters are identical to the previous section, except that we select the ``Step-Incremental'' strategy for the INT training. The resulting SIREN model has approximately 265K parameters.

\textbf{Megapixel Image} We fit the 8192$\times$8192 Pluto image \cite{pluto}. We use a SIREN model with 6 layers, each having 512 hidden units, resulting in a total of approximately 1M parameters. This model size is necessary to fit the image with 30+ PSNR. Model configurations are identical to that of 2D image fitting. We also train with Adam optimizer and cosine annealing learning rate scheduler, but instead, start with a learning rate of 1e-4. During training, we break the image into mini-batches of 524,288 pixels and have the INT algorithm sample training pixels for each optimization step. This is necessary due to VRAM constraints of consumer-grade GPUs such as NVIDIA RTX3090 (24GB). We train for a total of 500 epochs, where each epoch consists of 128 training steps that progressively sample the entire image. Thus, we also tune the incremental INT sampling interval to increase from 1 to 10 instead of from 1 to 100. 

\begin{minipage}{\linewidth}
  \begin{minipage}[b]{0.45\linewidth}
    \centering
    \includegraphics[width=\linewidth]{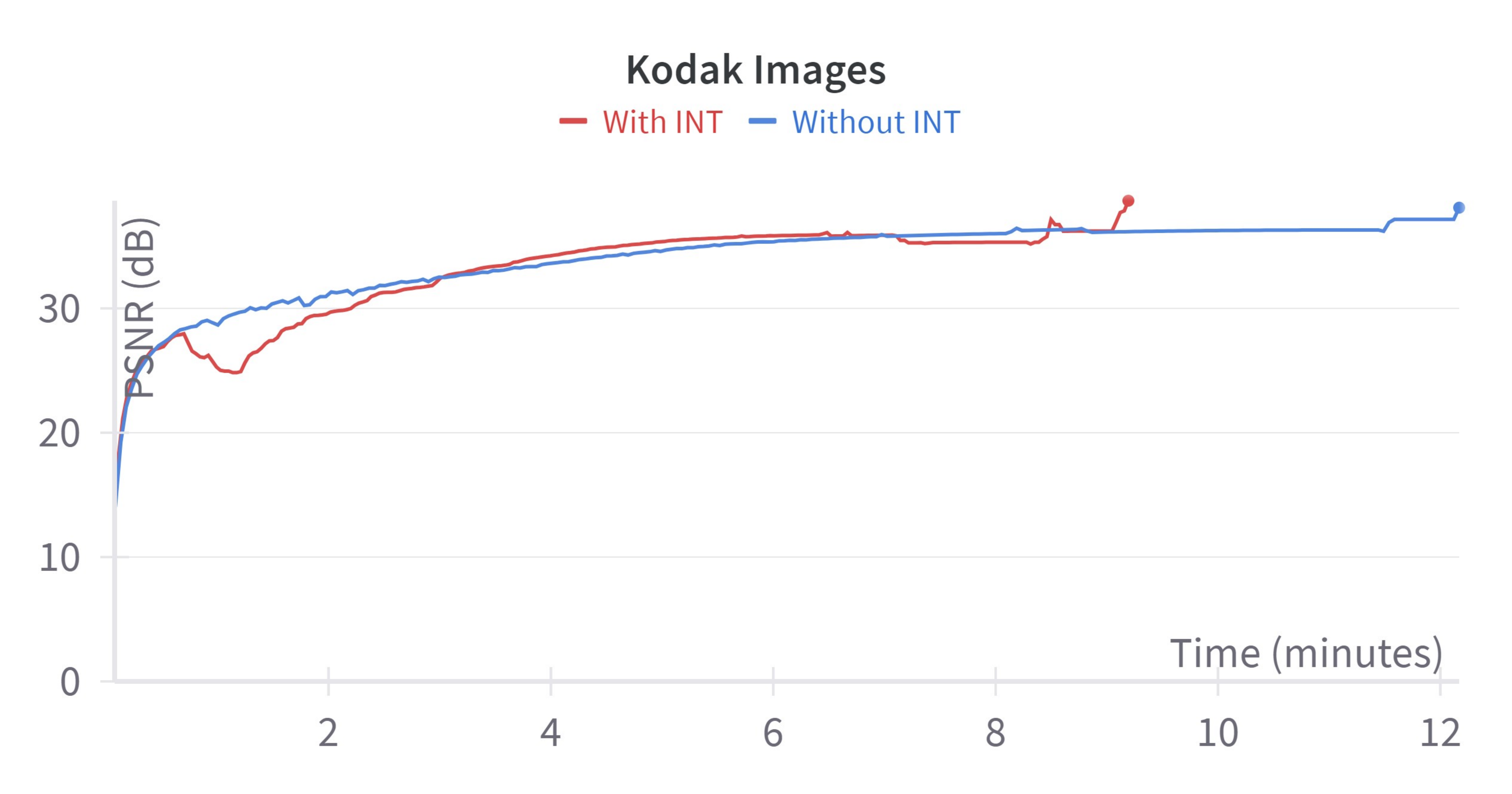}
    \captionof{figure}{PSNR-Training time curve of Kodak images training with and without INT.}
    \label{fig:kodak_psnr_curve}
  \end{minipage}
  \hfill
  \begin{minipage}[b]{0.45\linewidth}
    \includegraphics[width=\linewidth]{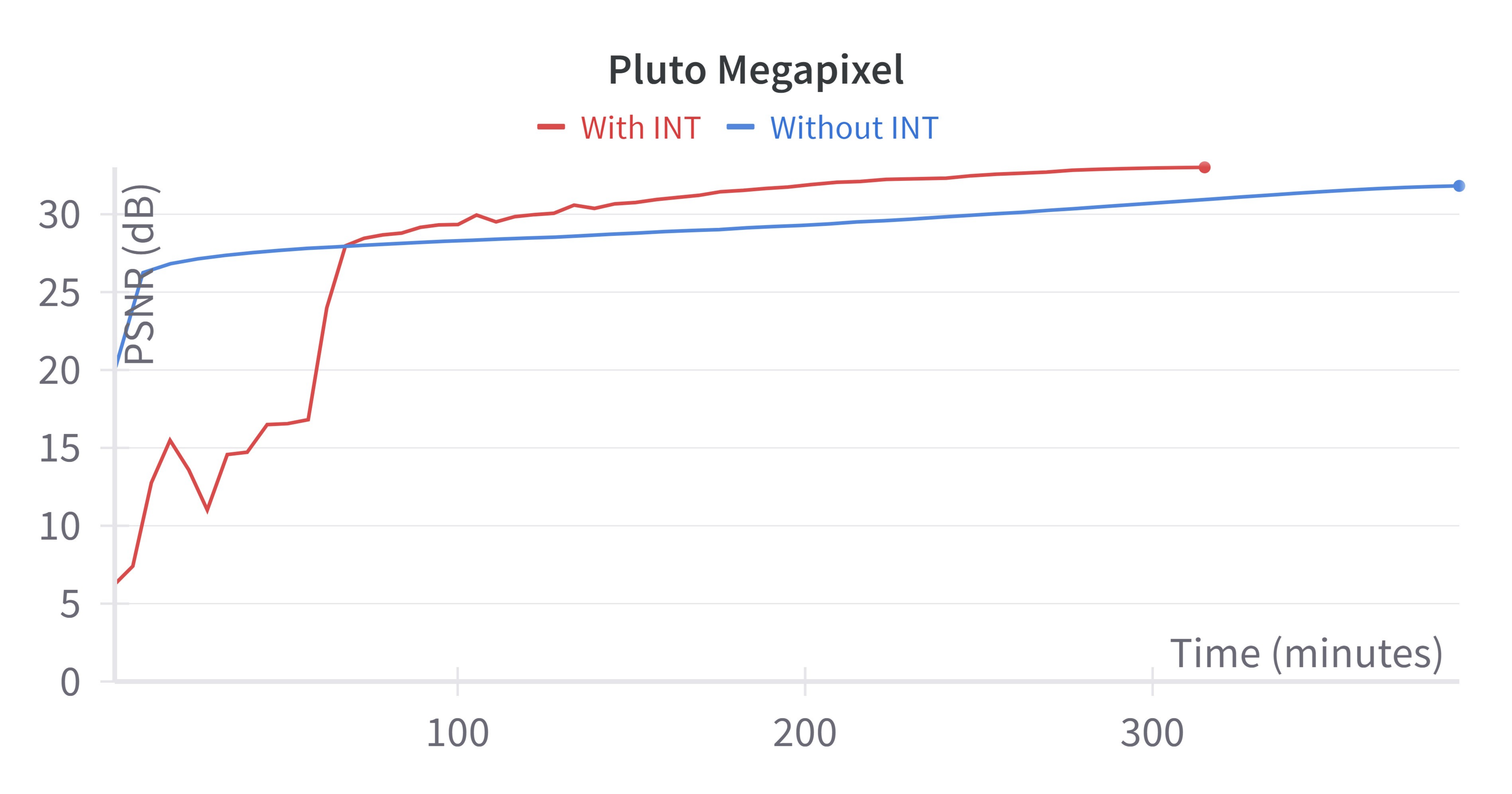}
    \captionof{figure}{PSNR-Training time curve of megapixel training with and without INT.}
    \label{fig:megapixel_psnr_curve}
  \end{minipage}
\end{minipage}

The ``non-smoothness" of training curve in Figure~\ref{fig:kodak_psnr_curve} and \ref{fig:megapixel_psnr_curve} is due to the increase in sampling intervals. In particular, the drop in reconstruction quality occurs when changing from densely selecting optimal training points at each iteration to sampling once per several iterations (as a measure of saving training time without sacrificing much ``final'' reconstruction quality). One can think of sampling at sparser intervals as analogous to training on dynamic minibatches of the data. Hence, at early stages of training when the model has not properly learnt the underlying signal yet, these minibatch training steps may lead to temporary overfitting and more ``jumpy'' training curves. However, our results show that this does not affect the ``final'' reconstruction quality. In fact, accompanying increasing INT ratio with sampling intervals is the optimal method of balancing lesser training samplers (faster training time) and retaining training quality.

\textbf{3D Shape.}  We conduct 3D shape experiments using the Stanford 3D Scanning Repository dataset~\cite{stanford3d}. We choose 4 scenes: Asian Dragon, Thai Statue, Lucy, and Armadillo. For our experiments, we utilize an 8-layer SIREN with 256 hidden units, resulting in approximately 400K parameters. Each scene is trained for 10K iterations. Following the approach of Bacon~\cite{lindell2022bacon} and Scone~\cite{li2023scone}, we sample points from the surface using a coarse and fine sampling procedure. We add two levels of Laplacian noise with variances of 1e-1 and 1e-3 for the coarse and fine samples, respectively. During each iteration, we randomly select a batch of 50K points. If INT is utilized, it is applied within each batch. IoU is computed by first transforming the learned signed distance function (SDF) to an occupancy grid of shape $512\times512\times512$ bounded by $[-0.5, 0.5]^3$. Below, we present the complete results for each scene:

\begin{table}[ht]
    \centering
    \begin{tabular}{c|cc}
         Scene & INT & IoU(\%) \\
         \toprule
         \multirow{2}{*}{Asian Dragon}  & \ding{55} & 96.46\\
                                        & \checkmark& 96.05 \\
         \midrule
         \multirow{2}{*}{Armadillo}  & \ding{55} & 98.48 \\
                                        & \checkmark& 98.25 \\
         \midrule
         \multirow{2}{*}{Thai Statue}  & \ding{55} & 96.43 \\
                                        & \checkmark& 96.22 \\
         \midrule
         \multirow{2}{*}{Lucy}  & \ding{55} & 96.91 \\
                                        & \checkmark& 96.19\\
        \bottomrule
    \end{tabular}
    \caption{3D shape representation results for all scenes.}
    \label{tab:3d_shape}
\end{table}

\end{document}